\documentclass[twoside, twocolumn]{article}

\usepackage[accepted]{aistats2025}

\usepackage{microtype}
\usepackage{graphicx}
\usepackage{subfigure}
\usepackage{booktabs} 
\usepackage{todonotes}
\usepackage{amsmath}
\usepackage{amsfonts}
\usepackage{soul}
\usepackage{subcaption}
\usepackage{natbib}

\usepackage{cancel}

\usepackage{fontenc}
\showoutput
\usetikzlibrary{fit}

\usepackage{hyperref}

\newcommand{\elbo}{\mathcal{L}}
\newcommand{\R}{\mathbb{R}}
\newcommand{\Xc}{\mathcal{H}}
\newcommand{\Sc}{\mathcal{F}}
\newcommand{\Rc}{\mathcal{R}}
\newcommand{\Dc}{\mathcal{Y}}

\begin{document}

%

%
\runningauthor{James Odgers, Ruby Sedgwick, Chrysoula Kappatou, Ruth Misener, Sarah Filippi}

\twocolumn[
\aistatstitle{Weighted-Sum of Gaussian Process Latent Variable Models}
\aistatsauthor{ James Odgers  \And Ruby Sedgwick \And  Chrysoula Kappatou }
\aistatsaddress{ Imperial College London \And Imperial College London \And Imperial College London } 
\aistatsauthor{ Ruth Misener \And Sarah Filippi }
\aistatsaddress{ Imperial College London \And Imperial College London } 
]

\begin{abstract}
This work develops a Bayesian non-parametric approach to signal separation where the signals may vary according to latent variables. 
Our key contribution is to augment Gaussian Process Latent Variable Models (GPLVMs) for the case where each data point comprises the weighted sum of a known number of pure component signals, observed across several input locations.
Our framework allows arbitrary non-linear variations in the signals while being able to incorporate useful priors for the linear weights, such as summing-to-one. 
Our contributions are particularly relevant to spectroscopy, where changing conditions may cause the underlying pure component signals to vary from sample to sample.
To demonstrate the applicability to both spectroscopy and other domains, we consider several applications: a near-infrared spectroscopy dataset with varying temperatures, a simulated dataset for identifying flow configuration through a pipe, and a dataset for determining the type of rock from its reflectance.
\end{abstract}

\section{Introduction}

\begin{figure}[t]
    \centering
    \begin{tikzpicture}
  \definecolor{mygray}{RGB}{169,169,169}

    \node[circle, fill = mygray, draw, minimum size=0.9cm] (d) at (3.5, 0) { ${y_{ij}}$};
    \node[circle, draw, minimum size=0.9cm] (r) at (2, 1.1) {{$r_{ic}$}};
    \node[circle, draw, minimum size=0.9cm] (x) at (0.6, 1.1) {{$h_{i}$}};
    \node[circle, draw, minimum size=0.9cm] (s) at (2, 0) {{$f_{ijc}$}};
    
    \node (l) at (-0.5,0) {{$\lambda_j$}};

    \node (c) at (2,1.6)[above] {{\scriptsize $c = 1, ..., C $}};
    \node (i) at (3.7,1)[above] {{\scriptsize $i = 1, ..., N $}};
    \node (j) at (-0.75,-0.6)[above] {{\scriptsize $j = 1, ..., M $}};

    \node[draw, fit=(r)(s)(c), inner sep=2pt, rounded corners=10] (box) {};
    \node[draw, fit=(r)(s)(d)(x)(i), inner sep=4pt, rounded corners=10] (box) {};
    \node[draw, fit=(j)(l)(s)(d), inner sep=3pt, rounded corners=10] (box) {};

    \draw[->] (l) -- (s);
    \draw[->] (x) -- (s);
    \draw[->] (r) -- (d);
    \draw[->] (s) -- (d);
\end{tikzpicture}
    \caption{Bayesian diagram for Weighted-Sum of Gaussian Process latent variable models (WS-GPLVM) with unobserved mixture weights $r_{ic}$. }

    \label{fig:diagram}
\end{figure}

\begin{figure*}[t]
\centering\begin{tikzpicture}
\node[inner sep=0pt, anchor=north west] (training_data) at (0,0)
    {\includegraphics[width=.3\textwidth]{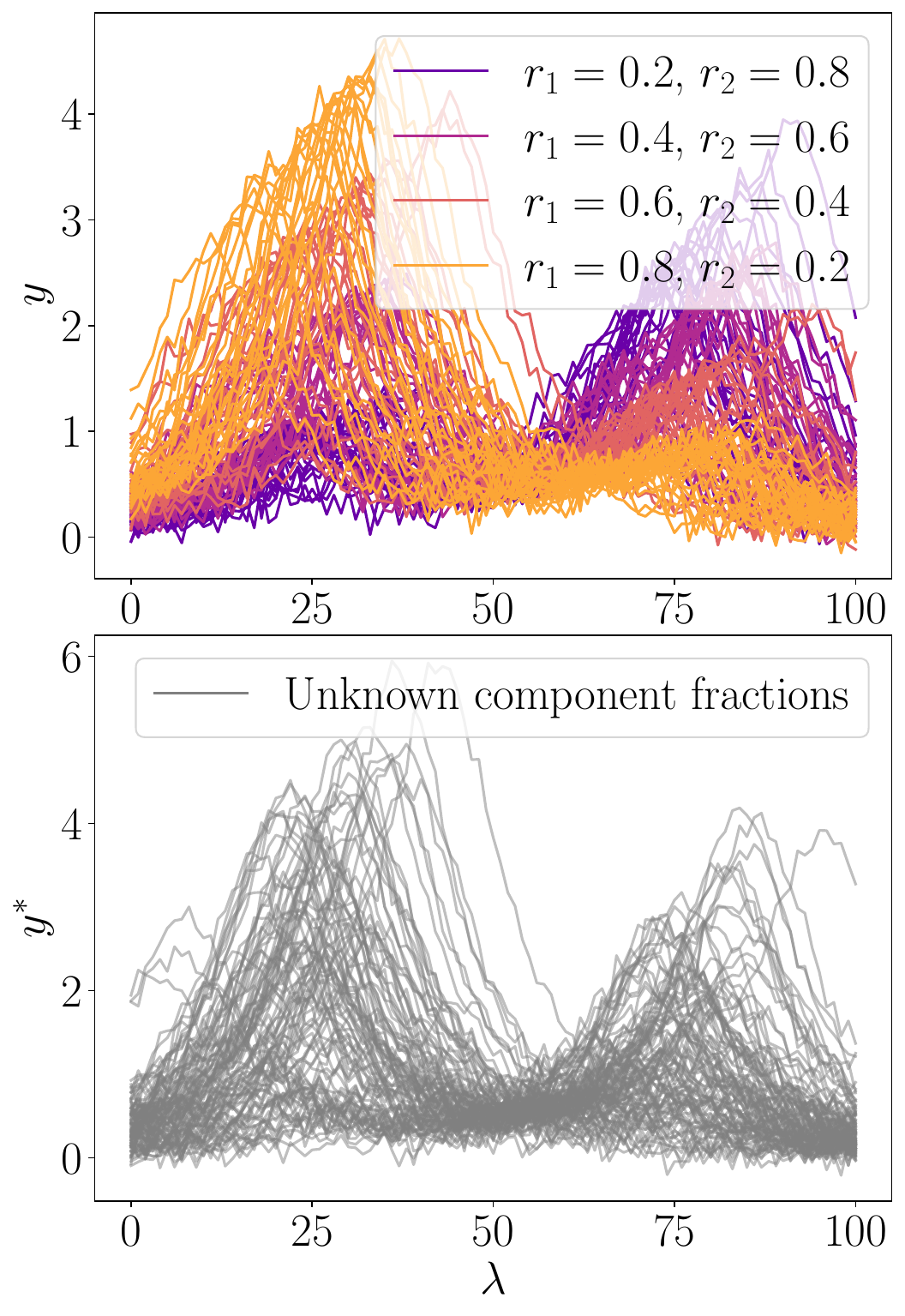}};

\node[inner sep=0pt, anchor = north west] (x) at (5.5,0)
    {\includegraphics[width=.32\textwidth]{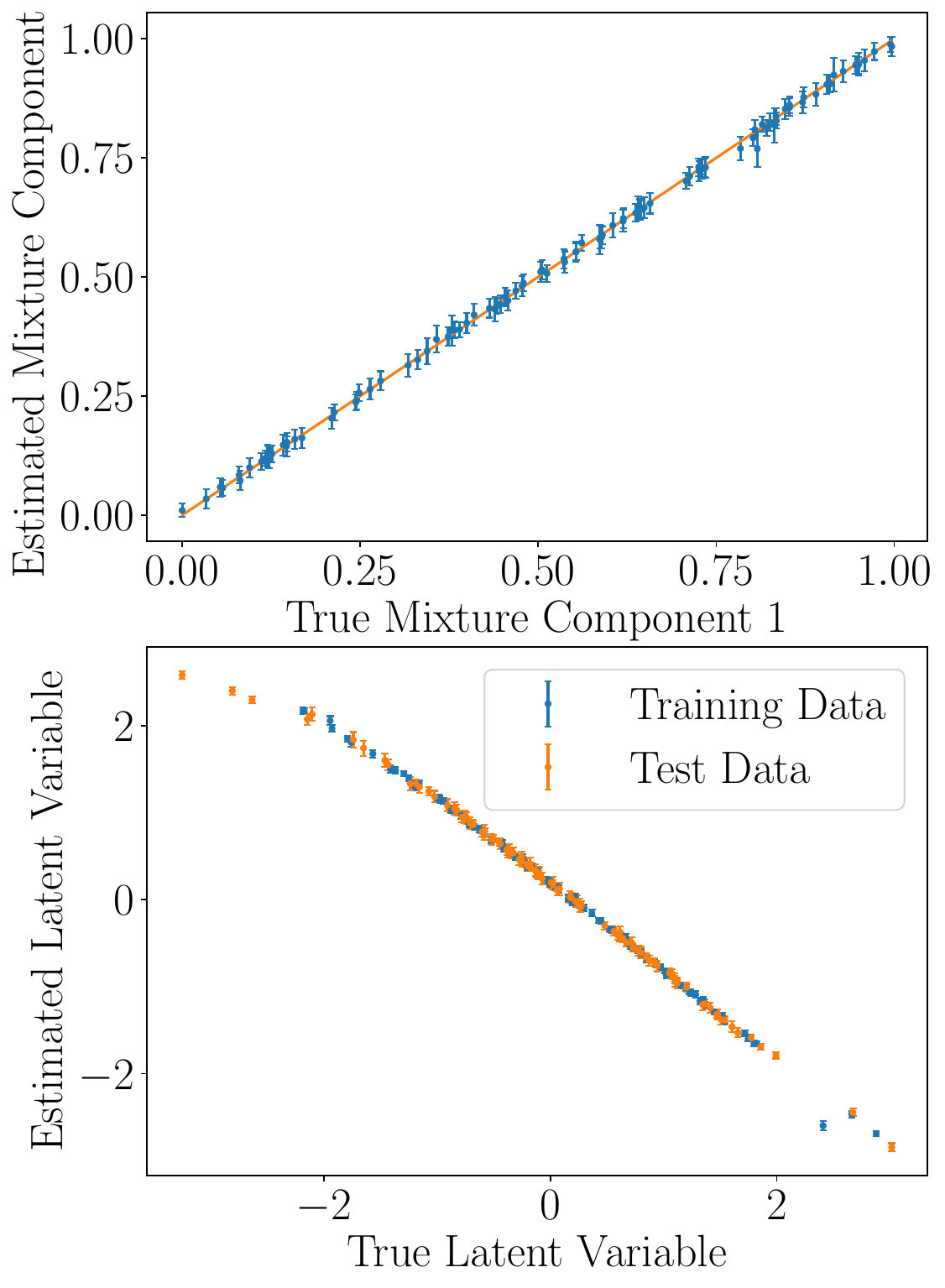}};
\node[inner sep=0pt, anchor = north west] (s) at (11,0)
    {\includegraphics[width=.3\textwidth]{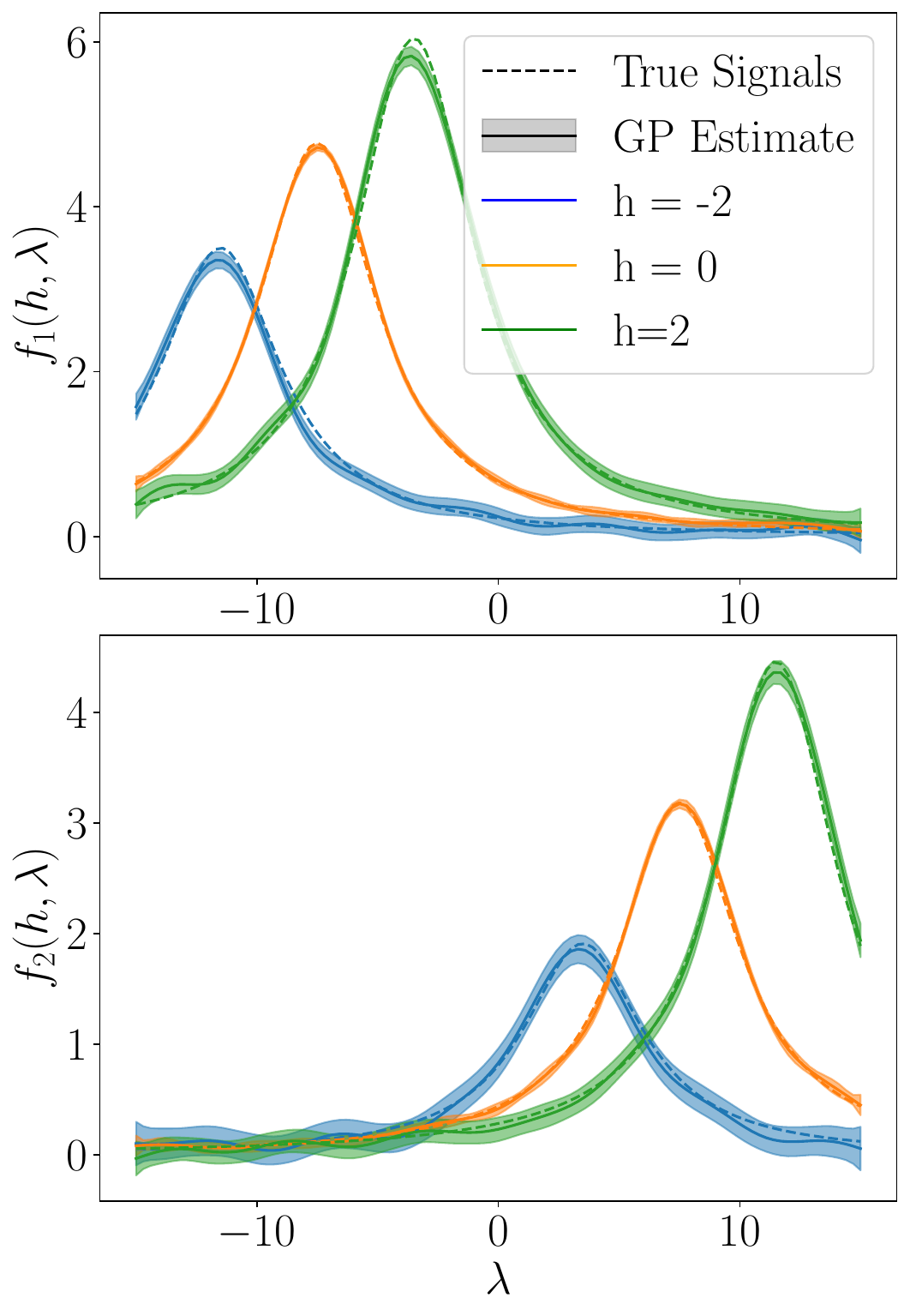}};

      \node[draw, fit=(training_data), inner sep=3pt, rounded corners=10, label = {[align = center] below: $(a)$ Observed data}] (box) {};
      \node[draw, fit=(x)(s), inner sep=3pt, rounded corners=10, label = {[align = center] below: $(b)$ Variational Posterior Estimates}] (box) {};

\end{tikzpicture}
\caption{Illustrative example of WS-GPLVM: by inputting a set of data $(a)$ of training data \textit{(top)}, with known mixture fractions, and test data \textit{(bottom)}, with unknown mixture fractions,  WS-GPLVM produces variational posterior estimates of $(b)$ the fractional composition of the test data \textit{(top left)}, the latent variables of both the training and the test data \textit{(bottom left)},  and a Gaussian Process estimate, shown with a mean and $95\%$ confidence interval, for the pure component signals \textit{(right)}.}

\label{fig: toy}
\end{figure*}

We develop a Bayesian non-parametric approach for handling additive models, focusing on \emph{signal separation}. Signal separation considers applications where a single observation is composed of several components and typically aims to either identify what are the underlying signals \citep{jayaram2020source, Chatterjee_2021_ICCV} or estimate the fraction of each signal present \citep{wytock2014contextually, yip2022esa}. 

{We motivate our approach by considering spectroscopy. In spectroscopy, for each data point $i$, a set of observations $\{y_{ij}| j =1,\ldots, M\}$ are made at a set of different wavelengths $\{\lambda_j | j=1,\dots, M\} $. Using these observations, we seek to identify the abundances of different chemical species $\{r_{ic}| c = 1, \ldots, C\}$ in the sample. 
Spectroscopic signals often naturally give rise to signal separation problems, due to Beer-Lambert's law, which states that light absorption at a particular wavelength due to a particular chemical species is proportional to the concentration of that chemical species and the path length through the medium. 
Adding complexity to the problem, the signal of each chemical, referred to in this paper as the \emph{pure component signals} ($\{f_c(\cdot, \cdot)| c = 1, \dots, C\}$), depends not only on the wavelength but also on other factors, for example, temperature \citep{wulfert1998influence}, which are often unobserved. 
Our key idea, and the starting point for this paper, is to use a latent variable $h_i \in \R^A$ to model these unobserved factors:
\begin{equation} \label{eq: data generation}
	 y_{ij} = \sum_{c=1}^C r_{ic}\; f_c(h_i, \lambda_j ) + \epsilon_{ij},
\end{equation}
where $\epsilon_{ij} \overset{iid}{\sim} N(0, \sigma^2)$ and the pure component signals $f_c(\cdot,\cdot)$ are unknown.} This is often combined with the additional constrain that the weights sum-to-one:
\begin{equation}
\label{eq:spectroscopy}
\sum_{c=1}^C r_{ic} = 1 \; \forall \; i \quad \text{and} \quad  r_{ic}\in[0,1] \; \forall \;i,c.
\end{equation}
Similar setups appear often in spectroscopy \citep{tauler1995multivariate,kriesten2008fully, munoz2020supervised},
but the formulation is more general and also used in other applications such as 
hyper-spectral unmixing \citep{altmann2013nonlinear,borsoi2019deep, borsoi2021spectral} and image/audio processing \citep{gandelsman2019double, jayaram2020source, mariani2023multi}. 


Our motivating example places constraints that no existing methods can solve. Namely, we are interested in settings where:
\begin{itemize}
    \item there is no parametric description of the signals, so we cannot rely on physical models \citep{alsmeyer2004indirect,dirks2019incorporating},
    \item there are relatively few observations of which only a small number are labeled, so we cannot rely on deep learning \citep{neri2021unsupervised, hong2021endmember}
    \item we may not have data on the pure signals $f_c$, that is we may lack data with all $r_{ic}$ equal 0 except one $\hat{c}$ with $r_{i\hat{c}} = 1$, so we cannot follow \cite{borsoi2019deep} and \cite{jayaram2020source}.
\end{itemize}

The challenges presented by these constraints necessitate appropriate regularization. 
Gaussian Processes (GPs) are an attractive choice for this due to their ability to automatically regularize through the optimization of the marginal likelihood \citep{williams2006gaussian}. 
As the conditions driving variability of the signals are unobserved, we use Gaussian Process Latent Variable Models (GPLVMs) \citep{lawrence2005probabilistic, titsias2010bayesian,JMLR:v17:damianou16a, dai2017efficient, de2021learning, ramchandran2021latent, lalchand2022generalised} to estimate what the conditions driving variation were. To the best of our knowledge, the setting where data arises from a weighted sum of GPLVMs has not been studied previously.

\textbf{Related Works. } Multi-output GPs often assume data is generated as the sum of unobserved latent functions, equivalent to what we refer to as pure component signals, with the goal of capturing covariance between different outputs
\citep{teh_semiparametric_2005, alvarez2011computationally, alvarez2008sparse, alvarez_kernels_2012}. 
While these methods can capture relationships between latent functions and observed data, none of them allow for variation in latent functions across different observations. By introducing latent variables as inputs to the latent functions our method captures variations in these latent functions which are not possible otherwise.

Gaussian Process Factor Analysis (GPFA) \citep{yu_gaussian-process_2008, luttinen_variational_2009, nadew2024conditionally} considers sums of GPs with the goal of extracting latent signals. Our method differs from this in two ways. Firstly, the GPFA model does not include the latent variable $h_i$, and secondly, their setting considered all weights to be unobserved whereas our setting has a mixture of both observed and unobserved weights. 

 Specifically for \textit{signal separation} settings, \citet{altmann2013nonlinear} use a GPLVM for hyper-spectral imaging 
 where the latent variables correspond to the unknown mixture fractional weights to capture a specific, physically motivated form of non-linearity.
In contrast, our approach incorporates latent variables which may be independent of the mixture fractional weights allowing us to capture a wider range of potentially unknown non-linearities in the pure component signals. 

\textbf{Contributions. } We propose a model to make predictions for data arising from Eq.~\eqref{eq: data generation} and \eqref{eq:spectroscopy} and derive an Evidence Lower Bound (ELBO) for this model to approximate the posterior distribution over the weights, pure component spectra and latent variables
, as described in Section \ref{sec:methods}. The optimization of this ELBO is challenging due to the model tending to rely on the non-linear variations of the GPLVM, so in Section~\ref{section: implimentation} we provide a training procedure which encourages the model to appropriately use both the latent variables, $h_i$,  and any unobserved weights, $r_i$. We demonstrate our approach on real datasets in Section~\ref{sec:experiments}.

\section{Weighted-Sum of Gaussian Process Latent Variable Models}\label{sec:methods}



We consider a dataset generated according to Eq.\ \ref{eq: data generation} and assume each data point $i = 1, \dots, N$ is observed at a fixed grid of $j = 1, \dots, M$ wavelengths.
We introduce the notation  $f_{ijc} = f_c(h_i, \lambda_j)\;$ for the pure component signals. Throughout this derivation, we need to aggregate across different indexes depending on whether we are working with the likelihood or the prior, for reasons discussed below. We denote this with a dot in the index over which we have aggregated into a vector, so $r_{i\cdot}$ is the vector containing all weights for data point $i$, and  $ f_{\cdot \cdot c}$ is the vector containing all of the pure component signals $c$ across all data points $i$ and wavelengths $j$, \textit{i.e.}   $r_{i\cdot} = (r_{i1}, \dots, r_{iC})^T$ and $f_{\cdot \cdot c} = (f_{11c}, \dots, f_{NMc})^T$. 

Spectroscopy applications typically have a set of training observations $Y=\{y_{ij}\}\in\R^{N\times M}$ with known weights $R=\{r_{ic}\}\in \R^{N\times C}$, and a set of test observations $Y^* = \{y^*_{ij}\} \in \R^{N^* \times M}$ with unknown weights. We aim to learn the posterior for the unknown weights of the test observations, $R^* = \{r^*_{ic}\} \in \R^{N^* \times C}$, the latent variables for the training and test data, denoted by $H=\{h_i\}\in  \R^{N \times A}$ and $H^*=\{h^*_i\}\in  \R^{N^* \times A}$ respectively, and the pure component signals, $F=\{f_{ijc}\}\in \R^{N\times M \times C}$ and $F^*=\{f^*_{ijc}\}\in \R^{N^*\times M \times C}$. 

Note that we use the labels training and test data to differentiate between data points where the weights are or are not observed. However, we assume that when fitting the model both the training and test data are available. In the model we define the posterior and variational parameters depend on the observed test spectra ($D^*$) as well as the training spectra and weights ($D, R$).

With this data, Eq.\ \eqref{eq: data generation} is non-identifiable. Even for data points with known weights, $r_{i\cdot}$, 
Eq.~\eqref{eq: data generation} defines a $C-1$ dimensional hyperplane of pure component signals, $f_{ij\cdot}$, which have equal likelihood. As there is no shared variable between each of the $N$ data points, more observations do not help. 
One solution is to put a prior on the pure component signals that creates dependence between $f_c(h_i, \lambda_j)$ and $f_c(h_{i'}, \lambda_{j'})$. 
%
%
%
In spectroscopy the degree of dependence between the data points will vary:  for data points with similar latent conditions, \textit{e.g.} data collected at similar temperatures and humidity, the model should share lots of information, \textit{i.e.} $f_c(h_i, \lambda_j) \approx f_c(h_{i'}, \lambda_{j})$ when $h_i \approx h_{i'}$. Meanwhile, data points collected under very different conditions may share very little information. Additionally, spectroscopy datasets are often small and can exhibit a wide range of possible variations. 
%
GPLVMs \citep{lawrence2005probabilistic, titsias2010bayesian} are an attractive prior in this context as they can learn the degree of dependence between $f_c(h_i, \lambda_j)$ and $f_c(h_{i'}, \lambda_{j'})$, while still being both flexible and data efficient enough to work on realistic datasets. 

Formally, we place a GP prior over each of the functions  $f_c(\cdot, \cdot)$, $c=1,\dots, C$, such that the vector $f_{\cdot \cdot c}$, containing all of the pure component signal values of that component, is distributed by
\begin{equation} 
    f_{\cdot \cdot c} \sim N(0, K^c_{\bullet \bullet}), \label{eq: s prior}
\end{equation}
where the kernel matrix is given by
\begin{equation}
    [K^c_{\bullet \bullet}]_{(ij)(i'j')} = k_c((h_i, \lambda_j),(h_{i'}, \lambda_{j'})). \label{eq: kernel}
\end{equation}
This prior is not the only plausible one to achieve the dependence between data points and we discuss another choice of GP prior in Section~\ref{sec: extensions}.


 

We use a Bayesian treatment of the latent variable, along with an ARD kernel to achieve automatic determination of the effective dimensionality of the latent space, as demonstrated by \citet{titsias2010bayesian}. The prior for the latent variable $h_i$ is a unit normal prior on the latent space:
\begin{equation}
    h_i \sim N(0, I_A). 
\end{equation}
The prior for $r_{i\cdot}=\{r_{ic}\}_{c=1}^C$ depends on the problem setting. For our applications, we primarily use a Dirichlet prior to force the weights to sum-to-one and be positive:
\begin{equation} \label{eq: ss prior}
    r_{i\cdot} \sim Dir(1_c).
\end{equation}
{Note that there is no requirement of conjugacy of the model imposed here.} We discuss other options for priors for $r_{i\cdot}$ in Section~\ref{sec: extensions}.


GPs are generally trained by maximizing the marginal likelihood \citep{williams2006gaussian}. However, the Bayesian treatment of the latent variables, $H$, means the exact marginal likelihood is no longer tractable. We therefore follow the variational inference approach of \citet{titsias2010bayesian} to approximate a lower bound of the posterior distribution over the quantities of interest, $F$, $F^*$, $H$, $H^*$ and $R^*$. 

More precisely, for each component $c$, we introduce a set of inducing points $u_c \in \R^L$ drawn from the Gaussian Process at a set of locations $V_{c}\in \R^{L\times A \times C}$ so that
\begin{equation} \label{eq: u prior}
    u_c \sim  N(0, K^c_{V_c V_c}),
\end{equation}
where the elements of the kernel matrix $K^c_{V_c V_c}$ are given by the kernel functions between the inducing point input locations.
With these inducing points, the conditional distribution for each pure component signal is
\begin{equation} \label{eq: augmented GP prior}
    f_{\cdot \cdot c}|u_c \sim N( \mu^f_{\cdot \cdot c}, K^f_{\cdot \cdot c} ),
\end{equation}
where 
\begin{equation*}
    \mu^f_{\cdot \cdot c} = K^c_{\bullet V_c}(K^c_{V_c V_c})^{-1}u_c
\end{equation*}
\begin{equation*}
    K^f_{\cdot \cdot c} = K^c_{\bullet \bullet } -  K^c_{\bullet V_c} (K^c_{V_c V_c})^{-1} K^{c \; \; T }_{\bullet V_c}.
\end{equation*}


 We are interested in the posterior distribution
\begin{equation}
\begin{split}
	&p   (\Sc, \Xc, R^*, U| \Dc,   R) 
 \\   
 & \qquad =  \frac{p(\Dc| \Rc, \Sc) p(\Sc|\Xc, U) p(\Rc) p(\Xc) p(U)}{p(\Dc,   R) },
\end{split}
\end{equation}
where 
for conciseness, we used the following notations:  $\Sc = \{F, F^*\}$, $\Xc = \{H, H^*\}$,  $\Dc = \{Y, Y^*\}$,  $\Rc = \{R, R^*\}$,  and $U=\{u_c\} \in \R^{L \times C}$ the matrix containing the inducing points for all the components. 


We aim to approximate this posterior distribution with a variational distribution of the form
\begin{equation}
    q(\Sc, \Xc, R^*, U) = p(\Sc| \Xc, U) q(\Xc) q(U) q(R^*)
\end{equation}
which, similarly to the Bayesian GPLVM \citep{titsias2010bayesian}, allows a simplification and an analytic computation of the ELBO given in Eq.~\ref{eq:elbo}. We use a mean field assumption for both $q(\Xc)$ and $q(R^*)$ such that the distributions factorize over the observations:
\begin{equation}
    q(\Xc) = \prod_{i=1}^N  N(h_{i}| \mu^{h}_i, \Sigma^h_i ) \prod_{i=1}^{N^*} N(h^*_{i}| \mu^{h^*}_i, \Sigma^{h^*}_i )
\end{equation} 
with $\Sigma^h_i \in \R^{A\times A}$ and $\Sigma^{h^*}_i \in \R^{A\times A}$ being  diagonal matrices 
and 
\begin{equation}
    q(R^*) = \prod_{i=1}^{N^*} q (r_{i\cdot}^*| \alpha_i),
\end{equation}
where $q(r_{i\cdot}^*| \alpha_i)$ is a Dirichlet distribution with parameter $\alpha_i \in \R^C$.
 We make no assumption on the form of $q(U)$, although it is shown in Appendix~\ref{appendix: ELBO derivation} to have an optimal distribution which is normal.

The optimal variational parameters, $\{\mu_i^x, \Sigma_i^x\}_{i=1}^N$, $\{\mu_i^{h^*}, \Sigma_i^{h^*}, \alpha_i\}_{i=1}^{N^*}$ and $\{r^*_{i\cdot}\}_{i=1}^C$, the inducing point locations, $\{V_c\}_{c=1}^C$, the Gaussian Process kernel hyper-parameters, and $\sigma^2$,
can be jointly found
by maximizing the Evidence Lower Bound (ELBO):
\begin{equation}
\label{eq:elbo}
\begin{split}
    \elbo &=  \langle \log\left(p(\Dc|\Sc,\Rc)\right) \rangle_{q(\Sc, \Xc, R^*, U)} 
    \\ & - D_{KL}\left(q(U)|| p(U))\right) - D_{KL}\left(q(\Xc)|| p(\Xc))\right)  
    \\ & - D_{KL}\left(q(R^*)|| p(R^*) \right) + \log\left(p(R)\right),
\end{split}
\end{equation}
where we use the notation $\langle \;\cdot\;\rangle_{q(\cdot)}$ to denote the expectation  with respect to $q(\cdot)$. 

The challenging term is the expected log-likelihood term, which is given by: 
\begin{equation*}
    \begin{split}
        & \langle \log\left(p(\Dc|\Sc,\Rc)\right) \rangle_{q(\Sc, \Xc, R^*, U)}  
        \\ & =  \sum_{ij}  \Biggl[2 y_{ij} r_{i\cdot}^T \langle f_{ij\cdot} \rangle_{q(\Sc, \Xc, U) } -  \langle f_{ij\cdot}^T r_{i\cdot} r_{i\cdot}^T f_{ij\cdot} \rangle_{q(\Sc, \Xc, U)}\Biggr]
        \\ &+   \sum_{ij}  \Biggl [y^*_{ij} \langle r_{i\cdot}^{* \; T} \rangle_{q(R^*)} \langle f^*_{ij\cdot} \rangle_{q(F^*,H^*,U) }
        \\ &\quad \quad- \langle f_{ij\cdot}^T \langle r_{i\cdot}^* r_{i\cdot}^{* \; T} \rangle_{q(R^*)} f_{ij\cdot} \rangle_{q(F^*,H^*,U)  } \Biggr] + B,
    \end{split}
\end{equation*}
where $B$ contains all the terms which do not depend on the variational parameters.

The derivation can be continued by using the results that
\begin{align*}
   & \langle r_{i\cdot}^{* \; T} \rangle_{q(R^*) } = \mu^r_{i\cdot}  \\
   & \langle r_{i\cdot}^* r_{i\cdot}^{* \; T} \rangle_{q(R^*)} = \mu^r_{i\cdot} \mu^{r \; T}_{i\cdot} + \Sigma^r_{i\cdot}, 
\end{align*}
where $\mu^r_{i\cdot} $  and $\Sigma^r_{i\cdot}$ are the mean and covariance of $q(r^*_{i\cdot}| \alpha_i)$. 
Similarly, 
\begin{equation*}
\begin{split}
    \langle f_{ij\cdot} \rangle_{q(F| X, U ) }& = \mu^f_{ij\cdot}
   \\ \langle f_{ij\cdot}^T r_{i\cdot}^T r_{i\cdot}^T   f_{ij\cdot}  \rangle_{p(F| X, U ) }  & =  \mu^{f \; T}_{ij\cdot} \left(\mu^r_{i\cdot} \mu^{r}_{i\cdot} + \Sigma^r_{i\cdot}\right) \mu^{f\; T}_{ij\cdot} 
    \\ &  + Tr \left( K^f_{ij\cdot}\left(\mu^r_{i\cdot} \mu^{r \; T}_{i\cdot} + \Sigma^r_{i\cdot}\right) \right), 
\end{split}
\end{equation*}
where $\mu^f_{ij\cdot} \in \R^C$ and $K^f_{ij\cdot} \in \R^{C \times C}$ are the mean and covariance for the Gaussian Process conditional on $U$ for measurement $i$ at location $j$. As $f_{ij\cdot}$ is a vector made from concatenating $C$ independent Gaussian Processes, the expressions for $\mu^{f}_{ij\cdot}$  and $K^f_{ij\cdot}$ depend on all of the values of $u_c$ together. To write these expressions we use the vectorized form of $U$, referred to as $u \in \R^{LC}$ in this paper, which has the prior distribution: 
\begin{equation}
    u \sim N(0, K_{VV})
\end{equation}
where $K_{VV}$ is a ${LC\times LC}$ block diagonal matrix:
\begin{equation*}
    K_{VV} = \text{block}(
        K^1_{V_1 V_1}, \dots,  K^C_{V_C V_C} )
\end{equation*} 
The values of $f_{ij\cdot}$ and $u$ are jointly normally distributed the mean and variance of the function values conditioned on the values of the inducing points are 
\begin{equation}
    \mu^f_{ij\cdot} = K_{(h_i, \lambda_j)V} K_{VV}^{-1} u
\end{equation}
\begin{equation}
    K^f_{ij\cdot} = K_{(h_i, \lambda_j)(h_i, \lambda_j)} - K_{(h_i, \lambda_j)V} K_{VV} K_{(h_i, \lambda_j)V}^T, 
\end{equation}
where the covariance between $f_{ij\cdot}$ and $u$ is a $C\times LC$ block diagonal matrix with the expression
\begin{equation*}
    K_{(h_i, \lambda_j)V} = \text{block}(
        K^1_{(h_i, \lambda_j) V_1}, \dots K^C_{(h_i, \lambda_j) V_C}) 
\end{equation*} 
and the covariance of $f_{ij\cdot}$ is the $C \times C$ matrix
\begin{equation*}
\begin{split}
   & K_{(h_i, \lambda_j)(h_i, \lambda_j)} = 
   \\ & \text{block}\left(
        k_1\big((h_i, \lambda_j) (h_i, \lambda_j)\big), \dots, k_c\big((h_i, \lambda_j),(h_i, \lambda_j)\big)\right) .
\end{split}
\end{equation*}
Plugging in the optimal form of $q(U)$ (see Appendix~\ref{appendix: ELBO derivation}) it can be shown that the ELBO is: 
    \begin{equation}
    \begin{split} 
        \mathcal{L}   = &  \frac{1}{2\sigma^4} \xi_1^T  \left(\frac{1}{\sigma^2} \xi_2 + K_{VV}\right)^{-1}  \xi_1  + \frac{1}{2} \log( \det(K_{VV})) 
        \\ & - \frac{1}{2}\log\left(\det\left(\frac{1}{\sigma^2} \xi_2 + K_{VV}\right)\right) - \frac{1}{2 \sigma^2}  \xi_{0}
        \\ &  +  \frac{1}{2\sigma^2}   
        Tr\left(K^{-1}_{VV}\xi_{2}\right) -  \frac{NM}{2} \log(2\pi\sigma^2) 
        \\ &  - \frac{1}{2\sigma^2 }\left(\sum_{i=1}^N \sum_{j=1}^M y_{i j}^2 + \sum_{i=1}^{N^*} \sum_{j=1}^M y^{*\; 2}_{i j}\right)  + \log(p(R))
        \\ & - KL\left(q(R^*)||p(R^*)\right)  - KL\left(q(\Xc)||p(\Xc)\right) 
    \end{split}
\end{equation}
where the quantities $\xi_0\in\R$, $\xi_1\in\R^{C \times LC}$ and $\xi_2\in \R^{LC \times LC}$ are analogous to the quantities often referred to as $\psi_0$, $\psi_1$ and $\psi_2$ in the Bayesian GPLVM literature, but in our setting also depend on the weights $r_{i\cdot}$. These are given by:
\begin{equation}
\begin{split} 
\notag
    \xi_0 = &  \sum_{  i =1 }^{  N}\sum_{j =1 }^{M} \left \langle Tr\left( r_{  i \cdot} r_{  i \cdot}^T K_{(h_{  i} \lambda_j )(h_{  i} \lambda_j )}\right)\right \rangle_{q( H)} 
\\    &+ \sum_{ i =1 }^{ N^*}\sum_{j =1 }^{M} \left \langle Tr\left(\left(\mu^r_{ i \cdot}\mu^{r \; T}_{ i \cdot} + \Sigma^r_{i \cdot }\right)K_{(h_{ i} \lambda_j )(h_{ i} \lambda_j )}\right)\right \rangle_{q(H^*)}
    \\    \xi_1   = & \sum_{  i =1}^{  N} \sum_{j=1}^M   y_{  i j}r^T_{  i \cdot} \left \langle K_{(  h_i \lambda_j)V}  \right \rangle_{q(  H)}
    \\ 
& + \sum_{ i =1}^{ N^*} \sum_{j=1}^M  y_{ij}\mu^{r\;T}_{  i \cdot} \left \langle K_{( h^*_i \lambda_j)V } \right \rangle_{q(  H^*)}\\
\xi_2 = &    \sum_{  i = 1 }^{  N} \sum_{j=1}^M \left \langle K^T_{(h_{  i} \lambda_j)V } r_{  i \cdot} r_{  i \cdot}^T K_{(h_i \lambda_j)V } \right \rangle_{q(  H)}  
    \\ & + \sum_{i=1}^{N^*} \sum_{j=1}^M \left \langle K^T_{(h_{ i} \lambda_j )V} \left(\mu^r_{ i \cdot}\mu^{r \; T}_{ i \cdot} + \Sigma^r_{i \cdot}\right) K_{(h_{i} \lambda _j )V} \right \rangle_{q(H^*)}.
\end{split}
\end{equation}
{
In the degenerate case where there is  only one component present, i.e. if $r_i = 1 $ and $p(r^*_{i \cdot }) =  q(r^*_{i \cdot }) = \delta(1-r^*_{i \cdot})$ for all $i$, the expressions $\xi_1, \xi_2$ and $\xi_3$ collapse exactly to the values of $\psi_1, \psi_2$ and $\psi_3$ from the Bayesian GPLVM literature \citep{titsias2010bayesian}. Interestingly, the expressions $\xi_1, \xi_2$ and $\xi_3$ are functions of the observed test spectra $D^*$, emphasizing that the variational posterior depends on the test spectra $D^*$ despite the absence of the observed weights $R^*$. 
}

The computational complexity of this method is $\mathcal{O}(L^3C^3)$ due to the need to invert $\left(\frac{1}{\sigma^2} \xi_2 + K_{VV}\right)^{-1} $. This bound is designed for smaller data sets in which this is still feasible, for large data sets the inverse could be avoided by finding the distributions for $q(u)$ using stochastic gradient descent \citep{hensman2013gaussian, lalchand2022generalised}.

\subsection{Extensions}
\label{sec: extensions}
\textbf{Different distributions for the weights.}
{The choice of the form of the distributions $q(R^*)$ and $p(R^*)$ is flexible due to the decision to numerically optimize these parameters, rather than use an analytic approach such as Coordinate Ascent Variational Inference (CAVI) \citep{blei2017variational}.} The only requirements are that (i) the first and second moments for a vector set of test weights $r^*_{i\cdot}$  are well defined and (ii) that the KL divergence between  $q(R^*)$  and $p(R^*) $ can be calculated. 
A simple approach, followed in this paper, is to select a variational distribution from the same family as the prior distribution.
So far we have used the Dirichlet distribution but our method is also applicable to other settings, for example, where $r_{ic}\in\{0,1\}$. 
In this case, where only one signal is present in each observation, we suggest using a categorical distribution for the a prior and variational distribution. This is similar to the method presented in \cite{lazaro-gredilla_overlapping_2012}, but extended to include latent variables. One of the examples in Section~\ref{sec:experiments} demonstrates the MO-GPLVM in this setting.


\textbf{Independent wavelengths.} In Equation~\eqref{eq: s prior} we assume that the pure component signals are correlated across wavelengths, which is desirable in spectroscopy when the pure component signals are smooth. 
However, for other applications there may not be covariance between the $M$ measurement locations, or the nature of the covariance may be unknown \emph{a-priori}. 
In this case, each of the  measurement locations can be treated as independent, so 
we place an independent GP prior for each location $j=1,\dots M$, that share the same kernel, so that Equation~\eqref{eq: s prior} becomes:
 \begin{equation*}
     f_{\cdot jc} \overset{iid}{\sim} N(0, K^{c}_{HH})  \;  \text{where} \; [K^c_{HH}]_{i i'} = k_c(h_i, h_{i'}).
 \end{equation*}
We call this version of our model WS-GPLVM-ind. The derivation of the ELBO for this specific case is provided 
in Appendix~\ref{appendix: independent measurements ELBO}.

\section{Implementation}
\label{section: implimentation}

Code to reproduce the results in this paper is available in the supplementary material and will be open-sourced after publication (see Appendix~\ref{appendix: reproducability} for more details).

\textbf{Initialization.} 
Maximizing the ELBO is a highly non-convex optimization problem meaning the initialization is critical. The means of GPLVM latent variables are commonly initialized using principal component analysis (PCA) \citep{lawrence2005probabilistic}. We adapt this by using PCA on the reconstruction error of the training set if it is assumed the pure signals do not vary between observations. The least squares estimate for a static signal is then
\begin{equation}
    \hat F =  Y ^T R (R^T R)^{-1}
\end{equation}
leading to reconstruction error for the training dataset
\begin{equation}
    E = Y - R \hat F^T . 
\end{equation}
We perform PCA on $E$ to get the initializations for the latent variables of the training data $\{\mu^h_{i}\}_{i=1}^N$. The initial variance $\Sigma^h_i$ is initialized $I_A$ for all $i$. The variational parameters of the test data are initialized to their prior distribution, i.e.\ for every $i$,  $q(h_i) = p(h)$ and $q(r_i) = p(r)$.
The other GP hyper-parameters, \{$\beta_c$, $\gamma_c$, $\sigma^2_{f_c}\}_{c=1}^C$, are initialized with constants (Appendix~\ref{appendix: Implementation Details}).

\textbf{Optimization.} 
Optimizing all model parameters together resulted in the model not using the weights, instead relying only on the flexibility of the GPLVM. To overcome this we developed a three-step approach to encourage our model to use the weights appropriately. (1) We infer the WS-GPLVM hyper-parameters and training data latent variational parameters, $\mu^h_i$ and $\Sigma^h_i$. (2) We then increase the noise variance $\sigma^2$ to a large value and optimize the test data's variational parameters, $\alpha^r_i$, $\mu^{h^*}_i$ and $\Sigma^{h^*}_i$, while progressively reducing $\sigma^2$ to the value it achieved at the end of step 1. (3) Finally, all variational and WS-GPLVM hyper-parameters are jointly optimized. See Appendix~\ref{appendix: Implementation Details} for more details.

The optimization is non-convex, meaning gradient descent methods can get stuck in local optima. To mitigate this, we do at least 5 random restarts, selecting the one with the highest ELBO. Appendix~\ref{appendix: Implementation Details} contains details of the initializations and random restarts. 

\textbf{GP Kernel.} The experiments in Section \ref{sec:experiments} use a single ARD kernel for all components.
The choice of the ARD kernel was motivated for automatic latent dimensionality selection \citep{titsias2010bayesian}, as well as an analytic computation of $\xi_i, \; i = 0,1,2$. These expressions are in Appendix~\ref{Appendix: xi}.


\textbf{Data pre-processing.} 
For the real datasets, there are often interferences which are not caused by the mixture model.  Two common interferences are (i) an offset of the entire signal by a constant and (ii) a multiplicative factor scaling the entire data. To mitigate these effects, prior work has suggested pre-processing procedures \citep{rakthanmanon2013addressing, dau2019ucr, rato2019ss, gerretzen2015simple,kappatou2023optimization}. We incorporate a pre-processing step called Standard Normal Variate (SNV) \citep{barnes1989standard} or z-normalisation \citep{rakthanmanon2013addressing} before fitting any models. In SNV, the data is transformed to have mean zero and variance one.


\section{Experiments}\label{sec:experiments}

\begin{figure*}[t!]
  \centering
  \begin{tikzpicture}
    \node[anchor=north west, inner sep=0, label = {[align = center] below: $(a)$ Observed Data}] at (0,4.8) {\includegraphics[width=0.29\linewidth]{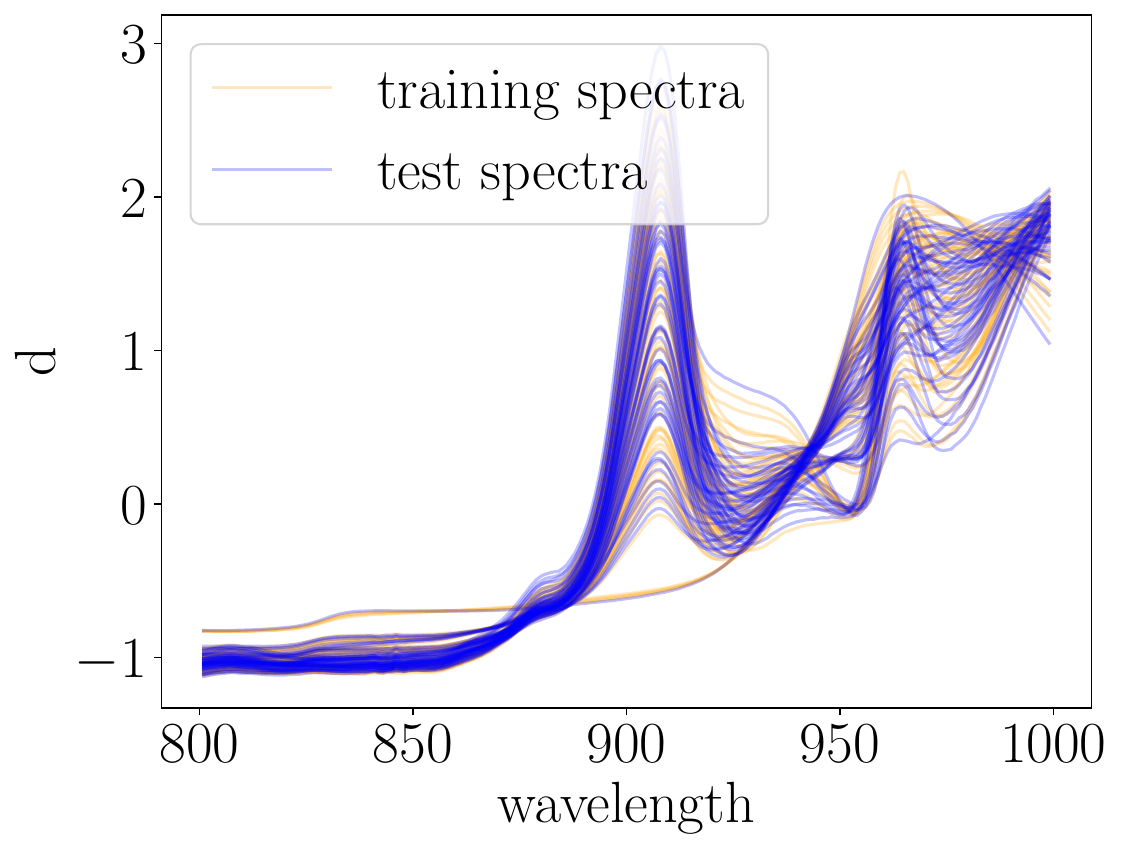}};

    \node[anchor=north west, inner sep=0 ,label = {[align = center] below: $(b)$ Estimated mixture fractions}] at (5,5.0) {\includegraphics[width=0.28\linewidth]{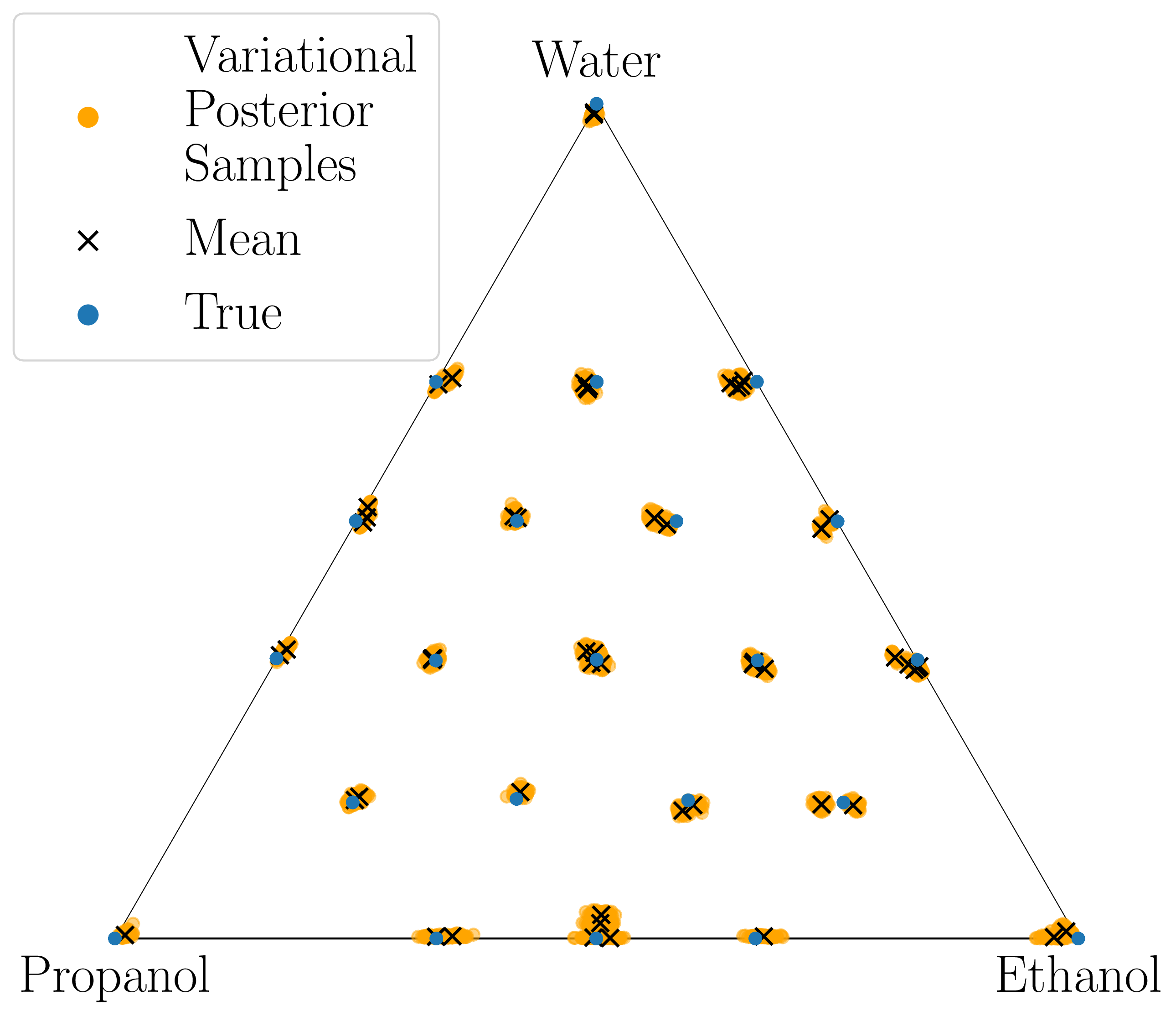}};
    \node at (0.5\linewidth, 1.5) {};

    \node[anchor=north west, inner sep=0, label = {[align = center] below: $(e)$ Mean  Pure Signals} ] at (0.6\linewidth,5.1) {\includegraphics[width=0.33\linewidth]{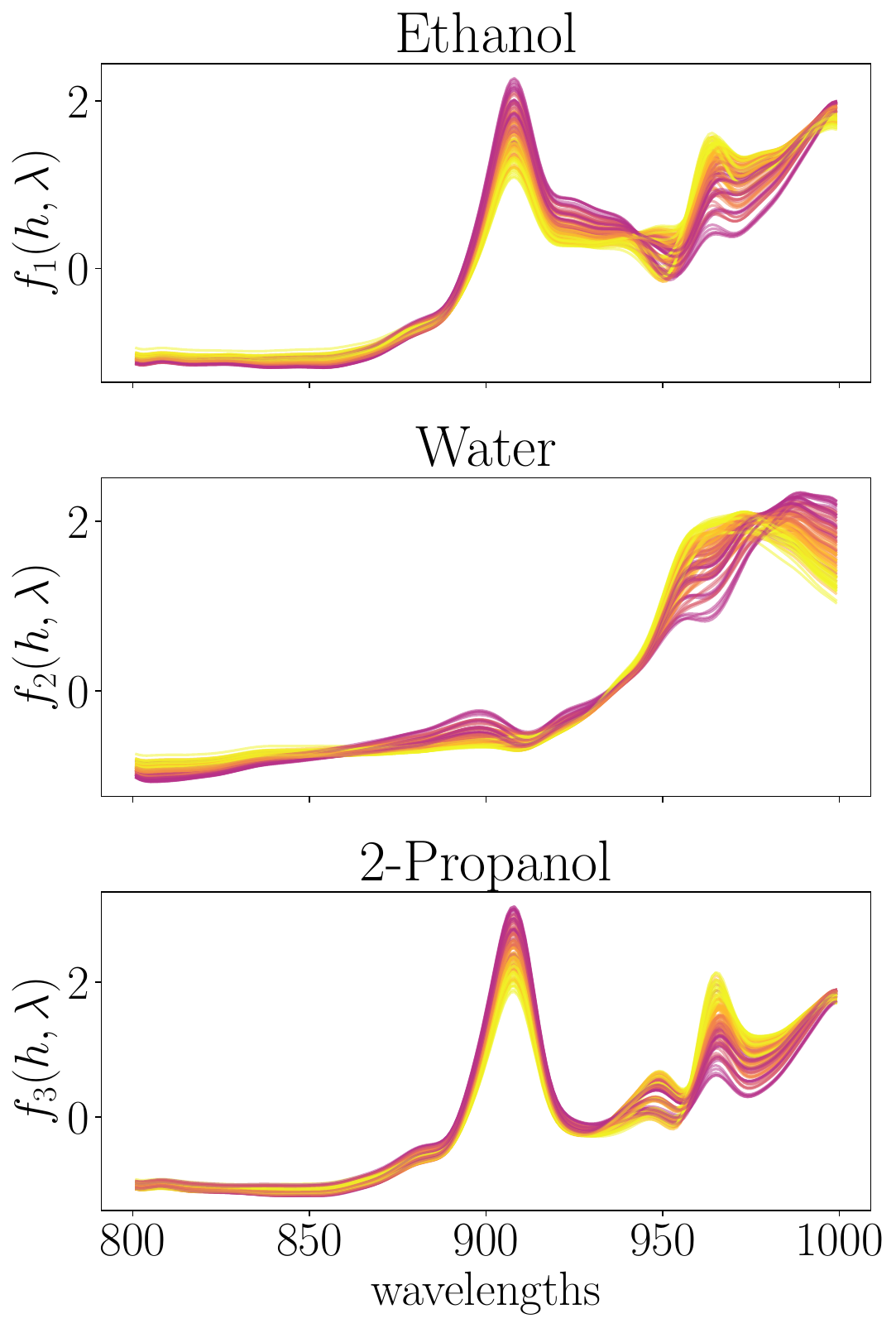}};
    \node at (0.85\linewidth, 1.5) {};


    \node[anchor=north west, inner sep=0,  label = {[align = center] below: $(c)$ $x$ colored by temperature}] at (0, 0.45) {\includegraphics[width=0.28\linewidth]{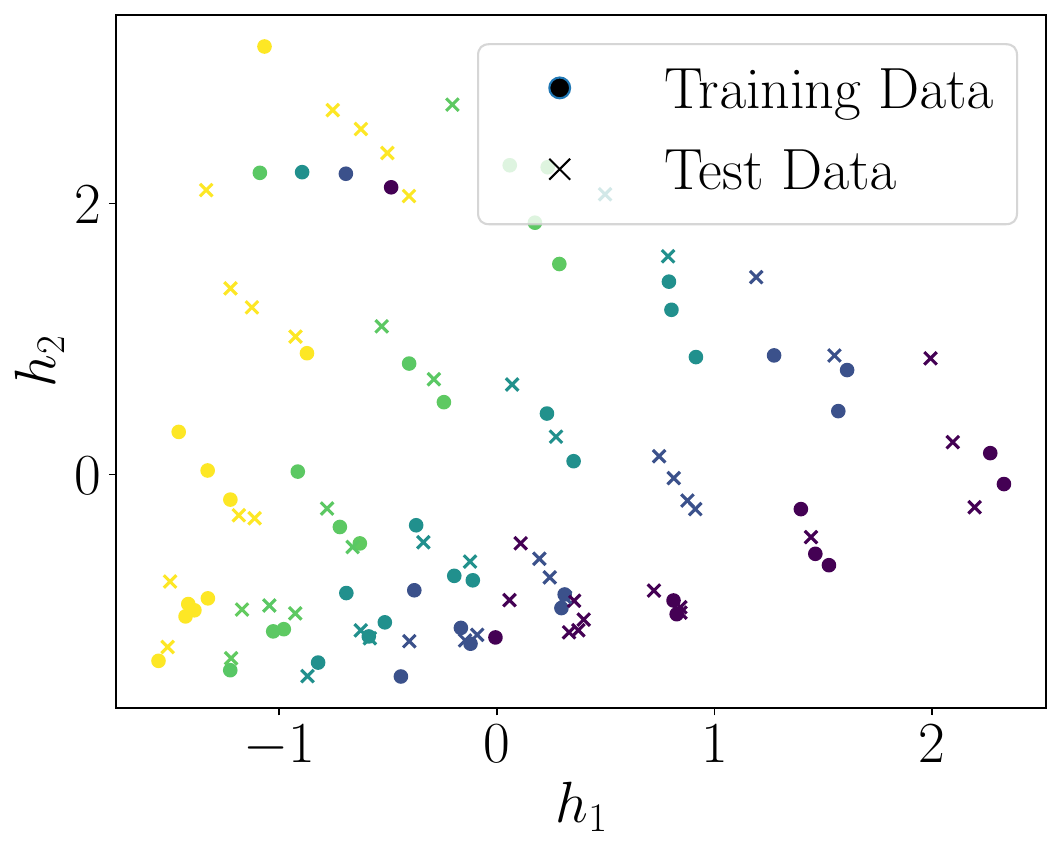}};

    \node[anchor=north west, inner sep=0,  label = {[align = center] below: $(d)$ $x$ colored by water fraction}] at (5, 0.45){\includegraphics[width=0.28\linewidth]{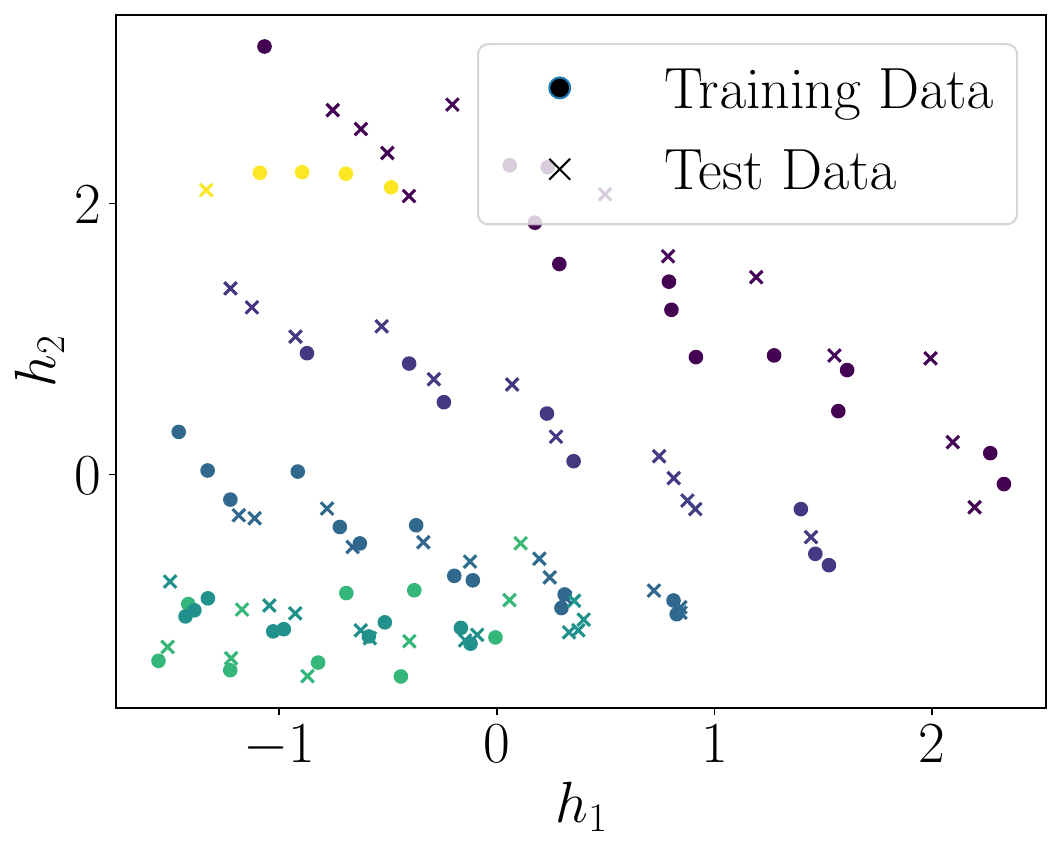}};
  \end{tikzpicture}
  \caption{\textit{(a)} Shows the training and test data after SNV was applied. Each orange line represents a training data point and each blue line is a test data point. \textit{(b)} Shows estimated mass fractions.  The blue points show the true values, the black crosses show the mean of our method's variational posterior, the orange points are 100 samples drawn from the variational posterior.  \textit{(c)} and \textit{(d)} show the latent variables colored by temperature (purple for $30^\circ C$, yellow for 70$^\circ C$) which we expected to cause variation in the pure signals, and water fraction (purple for no water, yellow for pure water), where the variation is induced by the normalizing procedure. \textit{(e)} Mean estimated pure component signals for the training and test spectra, evaluated at the mean of the learned latent variable and colored by the temperature of the measurement (red for $30^\circ C$, yellow for 70$^\circ C$).}
  \label{fig:wulfert figure}

\end{figure*}

\subsection{Toy Example}
Our toy example simulates two chemical species with distinct peaks which shift and scale depending on a single, one-dimensional latent variable. Figure \ref{fig: toy}$(a)$ illustrates the 100 training and 100 test observations with distinct peaks: the observations shift in both wavelength and change in height with changes to a single latent variable. We fit this data with a WS-GPLVM model with five latent dimensions, but the model correctly identified only one was relevant for predictions.
From the dataset illustrated in Figure \ref{fig: toy}$(a)$, we approximated the posterior of the pure signals ($f_c(\cdot, \cdot))$, latent variables ($x$) and unobserved mixture components ($r^*$). The Figure \ref{fig: toy}$(b)$ results demonstrate that our WS-GPLVM framework can effectively learn the values of all the unobserved variables when our assumptions hold. 

\subsection{Examples on Datasets from the Literature}

For the real-world examples, we compare to an inverse linear model of coregionalization model(ILMC), an approach based on Classical Least Squares (CLS/CLS-GP) and Partial Least Squares (PLS) \citep{wold2001pls, barker2003partial}. The CLS-GP model corresponds to  WS-GPLVM without the inclusion of the latent variables $h$. When the covariance across inputs is removed this model corresponds to CLS. 

Both ILMC and PLS are inverse models, \textit{i.e,} rather than treating the observed spectra $(Y)$ as an output and performing Bayesian inference, they treat the observed spectra as an input and attempt to directly predict the weights using $r*_{i\cdot} = g(y^*_{i\cdot}) + \eta_i$, where $g(\cdot)$ is a function to be learned from training data and $\eta_i$ is noise. PLS is a linear method which combines dimensionality reduction and prediction. It is a standard tool for making predictions from spectra. The ILMC uses a MOGP to estimate $g(\cdot)$, for this, we first reduce high dimensional inputs $(Y)$ using PCA, as GPs do not perform well with high dimensional inputs \citep{chen2007gaussian, binois2022survey}.  The dimensionality of the PCA embedding was selected using the ELBO. 

All baselines were run with five random restarts for each data split, selecting the one with the highest ELBO for ILMC and CLS/CLS-GP or cross-validation score for PLS. See Appendix~\ref{appendix:baseline_details} for full details. 


For regression, we analyze the models using Negative Log Predictive Density (NLPD) and the Root Mean Squared Error (RMSE) from the mean prediction to the true value of the test data. For classification, we compare predictive accuracy, Log Predictive Probability (LPP) and Area Under the Receiver Operator Curve (ROC AUC).   
Probabilistic predictions are not widely used in PLS \citep{odgers2023probabilistic}, so we do not report NLPD or LPP for PLS.

\textbf{Near Infra Red Spectroscopy with Varying Temperatures.} This dataset, from the chemometrics literature, explores the effects of changes in temperature on spectroscopic data \citep{wulfert1998influence}. The spectroscopic data, shown in Figure~\ref{fig:wulfert figure}, consists of a mixture of three components: water, ethanol and 2-propanol. These components were mixed in 22 different ratios, including pure components for each of them. Measurements were taken at 5 temperatures between $30^{\circ}C$ and $70^{\circ}C$. The temperature is not given to any of the models for making predictions. We pre-processed the data as described in Section~\ref{section: implimentation} and randomly split it into 10 training and test datasets, each with 50\% of the observations used for training and 50\% for testing, shown in Figure~\ref{fig:wulfert figure}$(a)$. 


Figures~\ref{fig:wulfert figure}$(b)$ and $(e)$ illustrate the performance of the proposed WS-GPLVM method in estimating the mixture fraction weights $(b)$ and the pure component signals $(e)$. 
The color of the estimated pure component signals indicates the temperature of the measurement: these variations are in agreement with the expected signals in \citet{wulfert1998influence}.

 Figure~\ref{fig:wulfert figure}$(c)$ and $(d)$ show the positions of the points in the two most significant latent variables found by the ARD kernel,
colored by the temperature of the sample (in Figure~\ref{fig:wulfert figure}$(c)$) or by the water fraction of the sample (in Figure~\ref{fig:wulfert figure}$(d)$). 
Latent variable dependence on the temperature was expected, but the latent variable depending on the water fraction was a surprise because water fraction is also available in the fractional weights $r_{ic}$.
We found this was due pre-processing that adjusted each spectrum to have a mean of zero and a variance of one, affecting samples differently based on their water mixture fractions. While this isn't directly modeled, the latent variables deal with this without interfering with the estimates for the mixture fractions. 

\begin{figure}[t]
    \centering
    \begin{minipage}{0.48\columnwidth}
        \includegraphics[width=\textwidth]{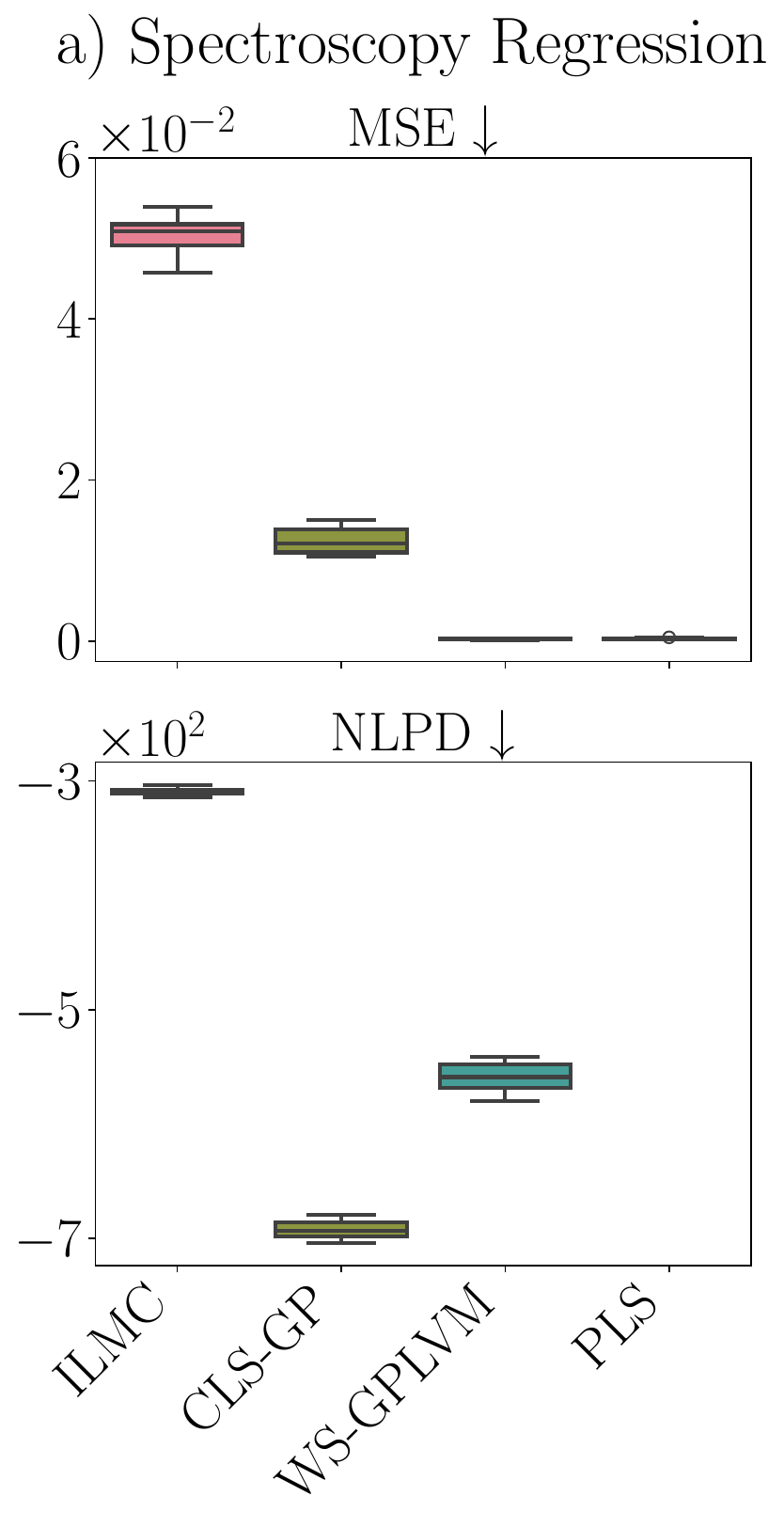}
    \end{minipage}%
    \hfill
    \begin{minipage}{0.48\columnwidth}
        \vspace{3mm}
        \includegraphics[width=\textwidth]{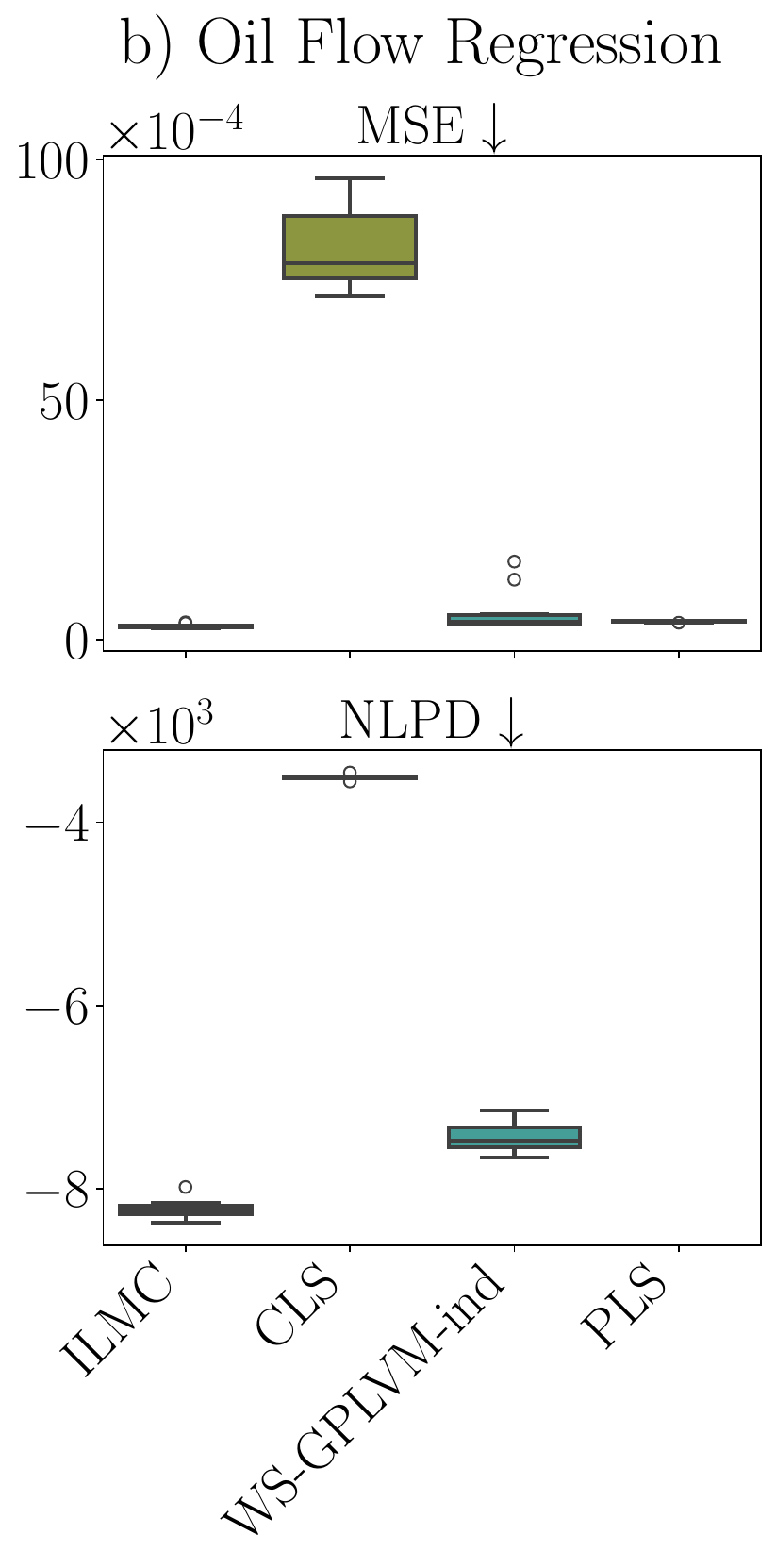}
    \end{minipage}
    \caption{Mean squared error (MSE) and negative log predictive density (NLPD) for 10 train-test splits of the a) near infra-red spectroscopy and b) oil flow datasets.}
    \label{fig:regression_results}
\end{figure}

 
For our comparison methods, the number of latent dimensions for PLS was selected based on a five-fold cross-validation of the training data. We used discrete Bayesian model selection to decide the number of PCA dimensions for the ILMC - for each training set eight GPs were fitted to PCA embeddings of dimensions $\{1..8\}$ and the one with the highest ELBO selected. 
Figure~\ref{fig:regression_results} shows the performance of WS-GPLVM and the baselines. WS-GPLVM had better mean accuracy with an MSE of $2.67\!\times\!10^{-4}$. The only comparable accuracy was  PLS with an MSE of $3.04\!\times\!10^{-4}$. The WS-GPLVM has a much better NLPD than the ILMC, but worse than CLS-GP. This indicates the WS-GPLVM may be overconfident, which is a known problem with variational inference \citep{blei2017variational}.

\textbf{Oil Flow Dataset.} 
This mixture dataset consists of simulated spectroscopic data of a mixture of oil, gas and water flowing through a pipe in different configurations, measured at two wavelengths, three horizontal and three vertical locations \citep{bishop1993analysis}. This dataset has been used in the GPLVM community \citep{lawrence2005probabilistic, titsias2009variational, urtasun2007discriminative, lalchand2022generalised} to demonstrate clustering of the classes of flow. Using WS-GPLVM, we can directly estimate the mixture fraction. 

This example has limited information on the spatial correlation of the measurements, so we use the WS-GPVLM-ind. model and consider all measurement locations independent. 
Additionally, the relationship between the mass fraction and the set of twelve measurements is known to be non-linear in a more pronounced way than the previous example after applying SNV. This challenges the latent variable's ability to account for the changes in the pure component signals as the fractional composition changes.

For this example, we don't perform PCA first for the ILMC as the original dimensionality is relatively low. We use the CLS baseline without covariance across inputs as there is no clear covariance structure to use. 
The aim here is to predict the fractional weights of oil, gas and water. 
Figure~\ref{fig:regression_results} outlines the results.

In this experiment, the inverse models (PLS and ILMC) perform the best, likely due to the non-linear relationships between the weights and the spectra. However, the WS-GPLVM-ind very nearly matches the performance of the inverse models in terms of accuracy and outperforms CLS by the greatest margin of any of our experiments. This experiment demonstrates the ability of the latent variable $h$ to greatly increase the performance of signal separation models when model misspecification could be a concern.

\textbf{Remote Sensing Rock Classification.} To demonstrate how the WS-GPLVM can be used for classification, i.e. $r_{ic} \in \{0,1\}$, we include a four class spectroscopic classification example from the UCR time series database \citep{baldridge2009aster,dau2019ucr}. This dataset has 70 data points, on which we did a 10-fold cross-validation to compare the different methods. For this example, we use a unit vector prior as described in Section~\ref{sec: extensions}.
Five-fold cross-validation was used to select the number of PLS dimensions, and the ELBO was used to select the number of PCA dimensions for the ILMC, out of $\{1..8\}$. A softmax likelihood was used to produce a valid categorical prediction.

The challenge here was that the training procedure needed to be adapted to work with a small classification dataset. Each classification example does not provide information about any other class so, based on the training data, there is no reason for different classes coming from the same latent point to be similar, which is not ideal to optimize the test set. We therefore skipped the initial pre-training of the model on just the training dataset, as described in Section~\ref{section: implimentation}.

\begin{figure}[t]
    \centering
    \includegraphics[width=\linewidth]{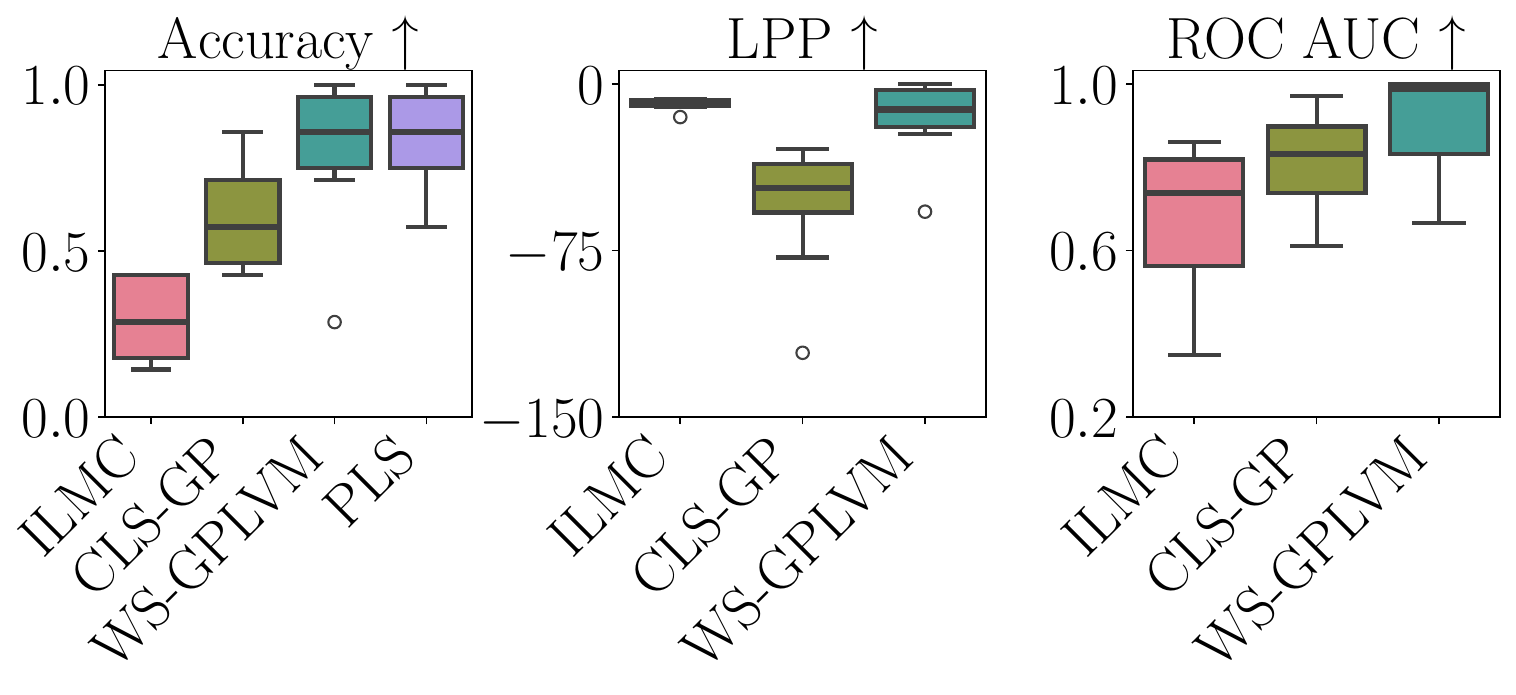}
    \caption{Accuracy, log predictive probability (LPP) and area under receiver operator curve (ROC AUC) for remote sensing rock classification for a 10-fold cross-validation.}
    \label{fig:spec_classification_results}
\end{figure}

Figure~\ref{fig:spec_classification_results} shows the results. For this dataset, WS-GPLVM and PLS are comparable, with mean accuracies of $81.4\%$ and $82.9\%$ and standard deviations of $21.3\%$ and $17.9\%$ respectively. The explicit estimate for the pure component signals in each class could be a compelling reason to choose WS-GPLVM over PLS. Note that while the ILMC method obtains a comparable to WS-GPLVM, it has a very low accuracy; this is due to the ILMC predicting roughly equal weights for all classes with a high level of uncertainty, therefore not making any meaningful classification.

\section{Discussion}
\label{sec:conclusion}
This paper extends GPLVMs from the output of a single GP to the case where we observe the sum of several GPs in different
mixture fractions. This framework is highly applicable to spectroscopic data where the summation in Eq.\ \eqref{eq: data generation} is a good approximation, but our approach is also effective in cases where the linear relationship between the mixture fractions and the observations does not hold exactly. We show that both regression and classification fit well within our model's framework and that our model complements existing GPLVM techniques.

\section*{Acknowledgments}
{Project funded by EPSRC Prosperity Grant (EP/T005556/1) in collaboration with Eli Lilly and Company, a BASF/RAEng Research Chair in Data-driven Optimisation to Ruth Misener, and by the Wellcome Trust [222836/Z/21/Z].}

\bibliography{references}

\begin{thebibliography}{57}
\providecommand{\natexlab}[1]{#1}
\providecommand{\url}[1]{\texttt{#1}}
\expandafter\ifx\csname urlstyle\endcsname\relax
  \providecommand{\doi}[1]{doi: #1}\else
  \providecommand{\doi}{doi: \begingroup \urlstyle{rm}\Url}\fi

\bibitem[Alsmeyer et~al.(2004)Alsmeyer, Ko{\ss}, and Marquardt]{alsmeyer2004indirect}
Frank Alsmeyer, Hans-J{\"u}rgen Ko{\ss}, and Wolfgang Marquardt.
\newblock Indirect spectral hard modeling for the analysis of reactive and interacting mixtures.
\newblock \emph{Applied spectroscopy}, 58\penalty0 (8):\penalty0 975--985, 2004.

\bibitem[Altmann et~al.(2013)Altmann, Dobigeon, McLaughlin, and Tourneret]{altmann2013nonlinear}
Yoann Altmann, Nicolas Dobigeon, Steve McLaughlin, and Jean-Yves Tourneret.
\newblock Nonlinear spectral unmixing of hyperspectral images using {Gaussian} processes.
\newblock \emph{IEEE Transactions on Signal Processing}, 61\penalty0 (10):\penalty0 2442--2453, 2013.
\newblock \doi{10.1109/TSP.2013.2245127}.

\bibitem[Alvarez and Lawrence(2008)]{alvarez2008sparse}
Mauricio Alvarez and Neil Lawrence.
\newblock Sparse convolved gaussian processes for multi-output regression.
\newblock \emph{Advances in neural information processing systems}, 21, 2008.

\bibitem[Alvarez and Lawrence(2011)]{alvarez2011computationally}
Mauricio~A Alvarez and Neil~D Lawrence.
\newblock Computationally efficient convolved multiple output gaussian processes.
\newblock \emph{The Journal of Machine Learning Research}, 12:\penalty0 1459--1500, 2011.

\bibitem[Baldridge et~al.(2009)Baldridge, Hook, Grove, and Rivera]{baldridge2009aster}
Alice~M Baldridge, Simon~J Hook, CI~Grove, and GJRSoE Rivera.
\newblock The {ASTER} spectral library version 2.0.
\newblock \emph{Remote sensing of environment}, 113\penalty0 (4):\penalty0 711--715, 2009.

\bibitem[Barker and Rayens(2003)]{barker2003partial}
Matthew Barker and William Rayens.
\newblock Partial least squares for discrimination.
\newblock \emph{Journal of Chemometrics: A Journal of the Chemometrics Society}, 17\penalty0 (3):\penalty0 166--173, 2003.

\bibitem[Barnes et~al.(1989)Barnes, Dhanoa, and Lister]{barnes1989standard}
RJ~Barnes, Mewa~Singh Dhanoa, and Susan~J Lister.
\newblock Standard normal variate transformation and de-trending of near-infrared diffuse reflectance spectra.
\newblock \emph{Applied spectroscopy}, 43\penalty0 (5):\penalty0 772--777, 1989.

\bibitem[Binois and Wycoff(2022)]{binois2022survey}
Mickael Binois and Nathan Wycoff.
\newblock A survey on high-dimensional gaussian process modeling with application to bayesian optimization.
\newblock \emph{ACM Transactions on Evolutionary Learning and Optimization}, 2\penalty0 (2):\penalty0 1--26, 2022.

\bibitem[Bishop and James(1993)]{bishop1993analysis}
Christopher~M Bishop and Gwilym~D James.
\newblock Analysis of multiphase flows using dual-energy gamma densitometry and neural networks.
\newblock \emph{Nuclear Instruments and Methods in Physics Research Section A: Accelerators, Spectrometers, Detectors and Associated Equipment}, 327\penalty0 (2-3):\penalty0 580--593, 1993.

\bibitem[Blei et~al.(2017)Blei, Kucukelbir, and McAuliffe]{blei2017variational}
David~M Blei, Alp Kucukelbir, and Jon~D McAuliffe.
\newblock Variational inference: A review for statisticians.
\newblock \emph{Journal of the American statistical Association}, 112\penalty0 (518):\penalty0 859--877, 2017.

\bibitem[Borsoi et~al.(2019)Borsoi, Imbiriba, and Bermudez]{borsoi2019deep}
Ricardo~Augusto Borsoi, Tales Imbiriba, and Jos{\'e} Carlos~Moreira Bermudez.
\newblock Deep generative endmember modeling: An application to unsupervised spectral unmixing.
\newblock \emph{IEEE Transactions on Computational Imaging}, 6:\penalty0 374--384, 2019.

\bibitem[Borsoi et~al.(2021)Borsoi, Imbiriba, Bermudez, Richard, Chanussot, Drumetz, Tourneret, Zare, and Jutten]{borsoi2021spectral}
Ricardo~Augusto Borsoi, Tales Imbiriba, Jos{\'e} Carlos~Moreira Bermudez, C{\'e}dric Richard, Jocelyn Chanussot, Lucas Drumetz, Jean-Yves Tourneret, Alina Zare, and Christian Jutten.
\newblock Spectral variability in hyperspectral data unmixing: A comprehensive review.
\newblock \emph{IEEE geoscience and remote sensing magazine}, 9\penalty0 (4):\penalty0 223--270, 2021.

\bibitem[Chatterjee et~al.(2021)Chatterjee, Le~Roux, Ahuja, and Cherian]{Chatterjee_2021_ICCV}
Moitreya Chatterjee, Jonathan Le~Roux, Narendra Ahuja, and Anoop Cherian.
\newblock Visual scene graphs for audio source separation.
\newblock In \emph{Proceedings of the IEEE/CVF International Conference on Computer Vision (ICCV)}, pages 1204--1213, 2021.

\bibitem[Chen et~al.(2007)Chen, Morris, and Martin]{chen2007gaussian}
Tao Chen, Julian Morris, and Elaine Martin.
\newblock Gaussian process regression for multivariate spectroscopic calibration.
\newblock \emph{Chemometrics and Intelligent Laboratory Systems}, 87\penalty0 (1):\penalty0 59--71, 2007.

\bibitem[Dai et~al.(2017)Dai, {\'A}lvarez, and Lawrence]{dai2017efficient}
Zhenwen Dai, Mauricio {\'A}lvarez, and Neil Lawrence.
\newblock Efficient modeling of latent information in supervised learning using {Gaussian} processes.
\newblock \emph{Advances in Neural Information Processing Systems}, 30, 2017.

\bibitem[Damianou et~al.(2016)Damianou, Titsias, and Lawrence]{JMLR:v17:damianou16a}
Andreas~C. Damianou, Michalis~K. Titsias, and Neil~D. Lawrence.
\newblock Variational inference for latent variables and uncertain inputs in {Gaussian} processes.
\newblock \emph{Journal of Machine Learning Research}, 17\penalty0 (42):\penalty0 1--62, 2016.

\bibitem[Dau et~al.(2019)Dau, Bagnall, Kamgar, Yeh, Zhu, Gharghabi, Ratanamahatana, and Keogh]{dau2019ucr}
Hoang~Anh Dau, Anthony Bagnall, Kaveh Kamgar, Chin-Chia~Michael Yeh, Yan Zhu, Shaghayegh Gharghabi, Chotirat~Ann Ratanamahatana, and Eamonn Keogh.
\newblock The {UCR} time series archive.
\newblock \emph{IEEE/CAA Journal of Automatica Sinica}, 6\penalty0 (6):\penalty0 1293--1305, 2019.

\bibitem[de~Souza et~al.(2021)de~Souza, Mesquita, Gomes, and Mattos]{de2021learning}
Daniel de~Souza, Diego Mesquita, Jo{\~a}o~Paulo Gomes, and C{\'e}sar~Lincoln Mattos.
\newblock Learning {GPLVM} with arbitrary kernels using the unscented transformation.
\newblock In \emph{International Conference on Artificial Intelligence and Statistics}, pages 451--459. PMLR, 2021.

\bibitem[Dirks and Poole(2019)]{dirks2019incorporating}
Matthew Dirks and David Poole.
\newblock Incorporating domain knowledge about xrf spectra into neural networks.
\newblock In \emph{Workshop on Perception as Generative Reasoning, NeurIPS}, 2019.

\bibitem[Gandelsman et~al.(2019)Gandelsman, Shocher, and Irani]{gandelsman2019double}
Yosef Gandelsman, Assaf Shocher, and Michal Irani.
\newblock ``double-dip": unsupervised image decomposition via coupled deep-image-priors.
\newblock In \emph{Proceedings of the IEEE/CVF conference on computer vision and pattern recognition}, pages 11026--11035, 2019.

\bibitem[Gardner et~al.(2018)Gardner, Pleiss, Weinberger, Bindel, and Wilson]{gardner2018gpytorch}
Jacob Gardner, Geoff Pleiss, Kilian~Q Weinberger, David Bindel, and Andrew~G Wilson.
\newblock Gpytorch: Blackbox matrix-matrix gaussian process inference with gpu acceleration.
\newblock \emph{Advances in neural information processing systems}, 31, 2018.

\bibitem[Gerretzen et~al.(2015)Gerretzen, Szyma\'nska, Jansen, Bart, van Manen, van~den Heuvel, and Buydens]{gerretzen2015simple}
Jan Gerretzen, Ewa Szyma\'nska, Jeroen~J Jansen, Jacob Bart, Henk-Jan van Manen, Edwin~R van~den Heuvel, and Lutgarde~MC Buydens.
\newblock Simple and effective way for data preprocessing selection based on design of experiments.
\newblock \emph{Analytical chemistry}, 87\penalty0 (24):\penalty0 12096--12103, 2015.

\bibitem[Hensman et~al.(2013)Hensman, Fusi, and Lawrence]{hensman2013gaussian}
James Hensman, Nicolo Fusi, and Neil~D Lawrence.
\newblock Gaussian processes for big data.
\newblock \emph{arXiv preprint arXiv:1309.6835}, 2013.

\bibitem[Hensman et~al.(2015)Hensman, Matthews, and Ghahramani]{hensman2015scalable}
James Hensman, Alexander Matthews, and Zoubin Ghahramani.
\newblock Scalable variational gaussian process classification.
\newblock In \emph{Artificial Intelligence and Statistics}, pages 351--360. PMLR, 2015.

\bibitem[Hong et~al.(2021)Hong, Gao, Yao, Yokoya, Chanussot, Heiden, and Zhang]{hong2021endmember}
Danfeng Hong, Lianru Gao, Jing Yao, Naoto Yokoya, Jocelyn Chanussot, Uta Heiden, and Bing Zhang.
\newblock Endmember-guided unmixing network ({EGU-Net}): A general deep learning framework for self-supervised hyperspectral unmixing.
\newblock \emph{IEEE Transactions on Neural Networks and Learning Systems}, 33\penalty0 (11):\penalty0 6518--6531, 2021.

\bibitem[Jayaram and Thickstun(2020)]{jayaram2020source}
Vivek Jayaram and John Thickstun.
\newblock Source separation with deep generative priors.
\newblock In \emph{International Conference on Machine Learning}, pages 4724--4735. PMLR, 2020.

\bibitem[Kappatou et~al.(2023)Kappatou, Odgers, Garc{\'\i}a-Mu{\~n}oz, and Misener]{kappatou2023optimization}
Chrysoula~D Kappatou, James Odgers, Salvador Garc{\'\i}a-Mu{\~n}oz, and Ruth Misener.
\newblock An optimization approach coupling preprocessing with model regression for enhanced chemometrics.
\newblock \emph{Industrial \& Engineering Chemistry Research}, 62\penalty0 (15):\penalty0 6196--6213, 2023.

\bibitem[King and Lawrence(2006)]{king2006fast}
Nathaniel~J King and Neil~D Lawrence.
\newblock Fast variational inference for gaussian process models through kl-correction.
\newblock In \emph{Machine Learning: ECML 2006: 17th European Conference on Machine Learning Berlin, Germany, September 18-22, 2006 Proceedings 17}, pages 270--281. Springer, 2006.

\bibitem[Kingma and Ba(2014)]{kingma2014adam}
Diederik~P Kingma and Jimmy Ba.
\newblock Adam: A method for stochastic optimization.
\newblock \emph{arXiv preprint arXiv:1412.6980}, 2014.

\bibitem[Kriesten et~al.(2008)Kriesten, Alsmeyer, Bardow, and Marquardt]{kriesten2008fully}
Ernesto Kriesten, Frank Alsmeyer, Andr{\'e} Bardow, and Wolfgang Marquardt.
\newblock Fully automated indirect hard modeling of mixture spectra.
\newblock \emph{Chemometrics and Intelligent Laboratory Systems}, 91\penalty0 (2):\penalty0 181--193, 2008.

\bibitem[Lalchand et~al.(2022)Lalchand, Ravuri, and Lawrence]{lalchand2022generalised}
Vidhi Lalchand, Aditya Ravuri, and Neil~D Lawrence.
\newblock Generalised gplvm with stochastic variational inference.
\newblock In \emph{International Conference on Artificial Intelligence and Statistics}, pages 7841--7864. PMLR, 2022.

\bibitem[Lawrence(2005)]{lawrence2005probabilistic}
Neil Lawrence.
\newblock Probabilistic non-linear principal component analysis with {Gaussian} process latent variable models.
\newblock \emph{Journal of machine learning research}, 6\penalty0 (11), 2005.

\bibitem[Luttinen and Ilin(2009)]{luttinen_variational_2009}
Jaakko Luttinen and Alexander Ilin.
\newblock Variational {Gaussian}-process factor analysis for modeling spatio-temporal data.
\newblock In \emph{Advances in {Neural} {Information} {Processing} {Systems}}, volume~22. Curran Associates, Inc., 2009.

\bibitem[Lázaro-Gredilla et~al.(2012)Lázaro-Gredilla, Van~Vaerenbergh, and Lawrence]{lazaro-gredilla_overlapping_2012}
Miguel Lázaro-Gredilla, Steven Van~Vaerenbergh, and Neil~D. Lawrence.
\newblock Overlapping {Mixtures} of {Gaussian} {Processes} for the data association problem.
\newblock \emph{Pattern Recognition}, 45\penalty0 (4):\penalty0 1386--1395, April 2012.
\newblock ISSN 0031-3203.
\newblock \doi{10.1016/j.patcog.2011.10.004}.

\bibitem[Mariani et~al.(2023)Mariani, Tallini, Postolache, Mancusi, Cosmo, and Rodol{\`a}]{mariani2023multi}
Giorgio Mariani, Irene Tallini, Emilian Postolache, Michele Mancusi, Luca Cosmo, and Emanuele Rodol{\`a}.
\newblock Multi-source diffusion models for simultaneous music generation and separation.
\newblock \emph{arXiv preprint arXiv:2302.02257}, 2023.

\bibitem[Mu\~noz and Torres(2020)]{munoz2020supervised}
Salvador~Garc\'ia Mu\~noz and Eduardo~Hern\'andez Torres.
\newblock Supervised extended iterative optimization technology for estimation of powder compositions in pharmaceutical applications: method and lifecycle management.
\newblock \emph{Industrial \& Engineering Chemistry Research}, 59\penalty0 (21):\penalty0 10072--10081, 2020.

\bibitem[Nadew et~al.(2024)Nadew, Fan, and Quinn]{nadew2024conditionally}
Yididiya~Y. Nadew, Xuhui Fan, and Christopher~John Quinn.
\newblock Conditionally-conjugate {G}aussian process factor analysis for spike count data via data augmentation.
\newblock In Ruslan Salakhutdinov, Zico Kolter, Katherine Heller, Adrian Weller, Nuria Oliver, Jonathan Scarlett, and Felix Berkenkamp, editors, \emph{Proceedings of the 41st International Conference on Machine Learning}, volume 235 of \emph{Proceedings of Machine Learning Research}, pages 37188--37212. PMLR, 21--27 Jul 2024.

\bibitem[Neri et~al.(2021)Neri, Badeau, and Depalle]{neri2021unsupervised}
Julian Neri, Roland Badeau, and Philippe Depalle.
\newblock Unsupervised blind source separation with variational auto-encoders.
\newblock In \emph{2021 29th European Signal Processing Conference (EUSIPCO)}, pages 311--315. IEEE, 2021.

\bibitem[Odgers et~al.(2023)Odgers, Kappatou, Misener, Garc{\'\i}a~Mu{\~n}oz, and Filippi]{odgers2023probabilistic}
James Odgers, Chrysoula Kappatou, Ruth Misener, Salvador Garc{\'\i}a~Mu{\~n}oz, and Sarah Filippi.
\newblock Probabilistic predictions for partial least squares using bootstrap.
\newblock \emph{AIChE Journal}, page e18071, 2023.

\bibitem[Paszke et~al.(2019)Paszke, Gross, Massa, Lerer, Bradbury, Chanan, Killeen, Lin, Gimelshein, Antiga, et~al.]{paszke2019pytorch}
Adam Paszke, Sam Gross, Francisco Massa, Adam Lerer, James Bradbury, Gregory Chanan, Trevor Killeen, Zeming Lin, Natalia Gimelshein, Luca Antiga, et~al.
\newblock Pytorch: An imperative style, high-performance deep learning library.
\newblock \emph{Advances in neural information processing systems}, 32, 2019.

\bibitem[Pedregosa et~al.(2011)Pedregosa, Varoquaux, Gramfort, Michel, Thirion, Grisel, Blondel, Prettenhofer, Weiss, Dubourg, et~al.]{pedregosa2011scikit}
Fabian Pedregosa, Ga{\"e}l Varoquaux, Alexandre Gramfort, Vincent Michel, Bertrand Thirion, Olivier Grisel, Mathieu Blondel, Peter Prettenhofer, Ron Weiss, Vincent Dubourg, et~al.
\newblock Scikit-learn: Machine learning in python.
\newblock \emph{the Journal of machine Learning research}, 12:\penalty0 2825--2830, 2011.

\bibitem[Rakthanmanon et~al.(2013)Rakthanmanon, Campana, Mueen, Batista, Westover, Zhu, Zakaria, and Keogh]{rakthanmanon2013addressing}
Thanawin Rakthanmanon, Bilson Campana, Abdullah Mueen, Gustavo Batista, Brandon Westover, Qiang Zhu, Jesin Zakaria, and Eamonn Keogh.
\newblock Addressing big data time series: Mining trillions of time series subsequences under dynamic time warping.
\newblock \emph{ACM Transactions on Knowledge Discovery from Data (TKDD)}, 7\penalty0 (3):\penalty0 1--31, 2013.

\bibitem[Ramchandran et~al.(2021)Ramchandran, Koskinen, and L{\"a}hdesm{\"a}ki]{ramchandran2021latent}
Siddharth Ramchandran, Miika Koskinen, and Harri L{\"a}hdesm{\"a}ki.
\newblock Latent {Gaussian} process with composite likelihoods and numerical quadrature.
\newblock In \emph{International Conference on Artificial Intelligence and Statistics}, pages 3718--3726. PMLR, 2021.

\bibitem[Rasmussen(2003)]{rasmussen2003gaussian}
Carl~Edward Rasmussen.
\newblock Gaussian processes in machine learning.
\newblock In \emph{Summer school on machine learning}. Springer, 2003.

\bibitem[Rato and Reis(2019)]{rato2019ss}
Tiago~J Rato and Marco~S Reis.
\newblock {SS-DAC}: A systematic framework for selecting the best modeling approach and pre-processing for spectroscopic data.
\newblock \emph{Computers \& Chemical Engineering}, 128:\penalty0 437--449, 2019.

\bibitem[Tauler(1995)]{tauler1995multivariate}
Roma Tauler.
\newblock Multivariate curve resolution applied to second order data.
\newblock \emph{Chemometrics and intelligent laboratory systems}, 30\penalty0 (1):\penalty0 133--146, 1995.

\bibitem[Teh et~al.(2005)Teh, Seeger, and Jordan]{teh_semiparametric_2005}
Yee~Whye Teh, Matthias Seeger, and Michael~I Jordan.
\newblock Semiparametric {Latent} {Factor} {Models}.
\newblock In \emph{International {Workshop} on {Artificial} {Intelligence} and {Statistics}}, pages 333--340, 2005.

\bibitem[Titsias(2009)]{titsias2009variational}
Michalis Titsias.
\newblock Variational learning of inducing variables in sparse {Gaussian} processes.
\newblock In \emph{Artificial intelligence and statistics}, pages 567--574. PMLR, 2009.

\bibitem[Titsias and Lawrence(2010)]{titsias2010bayesian}
Michalis Titsias and Neil~D Lawrence.
\newblock Bayesian {Gaussian} process latent variable model.
\newblock In \emph{Proceedings of the thirteenth international conference on artificial intelligence and statistics}, pages 844--851. JMLR Workshop and Conference Proceedings, 2010.

\bibitem[Urtasun and Darrell(2007)]{urtasun2007discriminative}
Raquel Urtasun and Trevor Darrell.
\newblock Discriminative {Gaussian} process latent variable model for classification.
\newblock In \emph{Proceedings of the 24th international conference on Machine learning}, pages 927--934, 2007.

\bibitem[Williams and Rasmussen(2006)]{williams2006gaussian}
Christopher~KI Williams and Carl~Edward Rasmussen.
\newblock \emph{Gaussian processes for machine learning}.
\newblock MIT press Cambridge, MA, 2006.

\bibitem[Wold et~al.(2001)Wold, Sj{\"o}str{\"o}m, and Eriksson]{wold2001pls}
Svante Wold, Michael Sj{\"o}str{\"o}m, and Lennart Eriksson.
\newblock {PLS}-regression: a basic tool of chemometrics.
\newblock \emph{Chemometrics and intelligent laboratory systems}, 58\penalty0 (2):\penalty0 109--130, 2001.

\bibitem[W{\"u}lfert et~al.(1998)W{\"u}lfert, Kok, and Smilde]{wulfert1998influence}
Florian W{\"u}lfert, Wim~Th Kok, and Age~K Smilde.
\newblock Influence of temperature on vibrational spectra and consequences for the predictive ability of multivariate models.
\newblock \emph{Analytical chemistry}, 70\penalty0 (9):\penalty0 1761--1767, 1998.

\bibitem[Wytock and Kolter(2014)]{wytock2014contextually}
Matt Wytock and J~Kolter.
\newblock Contextually supervised source separation with application to energy disaggregation.
\newblock In \emph{Proceedings of the AAAI Conference on Artificial Intelligence}, volume~28, 2014.

\bibitem[Yip et~al.(2022)Yip, Waldmann, Changeat, Morvan, Al-Refaie, Edwards, Nikolaou, Tsiaras, de~Oliveira, Lagage, et~al.]{yip2022esa}
Kai~Hou Yip, Ingo~P Waldmann, Quentin Changeat, Mario Morvan, Ahmed~F Al-Refaie, Billy Edwards, Nikolaos Nikolaou, Angelos Tsiaras, Catarina~Alves de~Oliveira, Pierre-Olivier Lagage, et~al.
\newblock {ESA-Ariel} data challenge {NeurIPS} 2022: Inferring physical properties of exoplanets from next-generation telescopes.
\newblock \emph{arXiv preprint arXiv:2206.14642}, 2022.

\bibitem[Yu et~al.(2008)Yu, Cunningham, Santhanam, Ryu, Shenoy, and Sahani]{yu_gaussian-process_2008}
Byron~M Yu, John~P Cunningham, Gopal Santhanam, Stephen Ryu, Krishna~V Shenoy, and Maneesh Sahani.
\newblock Gaussian-process factor analysis for low-dimensional single-trial analysis of neural population activity.
\newblock In \emph{Advances in {Neural} {Information} {Processing} {Systems}}, volume~21. Curran Associates, Inc., 2008.

\bibitem[Álvarez et~al.(2012)Álvarez, Rosasco, and Lawrence]{alvarez_kernels_2012}
Mauricio~A. Álvarez, Lorenzo Rosasco, and Neil~D. Lawrence.
\newblock Kernels for {Vector}-{Valued} {Functions}: {A} {Review}.
\newblock \emph{Foundations and Trends in Machine Learning}, 4\penalty0 (3):\penalty0 195--266, June 2012.
\newblock ISSN 1935-8237, 1935-8245.
\newblock \doi{10.1561/2200000036}.
\newblock Publisher: Now Publishers, Inc.

\end{thebibliography}
\bibliographystyle{plainnat}

\section*{Checklist}



 \begin{enumerate}

 \item For all models and algorithms presented, check if you include:
 \begin{enumerate}
   \item A clear description of the mathematical setting, assumptions, algorithm, and/or model. [Yes]
   We give a clear explanation in section 2. 
   \item An analysis of the properties and complexity (time, space, sample size) of any algorithm. [Yes]
   We provide this at the end of section 2. 
   \item (Optional) Anonymized source code, with specification of all dependencies, including external libraries. [Yes]
 The code is available.
 \end{enumerate}

 \item For any theoretical claim, check if you include:
 \begin{enumerate}
   \item Statements of the full set of assumptions of all theoretical results. [Not Applicable]
    We do not make theoretical claims about our models performance. 
   \item Complete proofs of all theoretical results. [Not Applicable]
    We do not make theoretical claims about our models performance. 
   \item Clear explanations of any assumptions. [Yes]   
  We are clear about our assumptions throughout. 
 \end{enumerate}

 \item For all figures and tables that present empirical results, check if you include:
 \begin{enumerate}
   \item The code, data, and instructions needed to reproduce the main experimental results (either in the supplemental material or as a URL). [Yes]
   The code is available. 
   \item All the training details (e.g., data splits, hyperparameters, how they were chosen). [Yes]
   Details in Section~\ref{sec:experiments} and Appendix~\ref{appendix: Implementation Details}. 
    \item A clear definition of the specific measure or statistics and error bars (e.g., with respect to the random seed after running experiments multiple times). [Yes]   All results are reported with error bars. 
    \item A description of the computing infrastructure used. (e.g., type of GPUs, internal cluster, or cloud provider). [Yes]
   See Appendix~\ref{appendix: reproducability}
 \end{enumerate}

 \item If you are using existing assets (e.g., code, data, models) or curating/releasing new assets, check if you include:
 \begin{enumerate}
   \item Citations of the creator If your work uses existing assets. [Yes]
 See Appendix~\ref{appendix: reproducability}.
   \item The license information of the assets, if applicable. [Yes]
  ee Appendix~\ref{appendix: reproducability}.
   \item New assets either in the supplemental material or as a URL, if applicable. [Not Applicable]
  
   \item Information about consent from data providers/curators. [Not Applicable]
  
   \item Discussion of sensible content if applicable, e.g., personally identifiable information or offensive content. [Not Applicable]
   
 \end{enumerate}

 \item If you used crowdsourcing or conducted research with human subjects, check if you include:
 \begin{enumerate}
   \item The full text of instructions given to participants and screenshots. [Not Applicable]
   \item Descriptions of potential participant risks, with links to Institutional Review Board (IRB) approvals if applicable. [Not Applicable]

   \item The estimated hourly wage paid to participants and the total amount spent on participant compensation. [Not Applicable] 

 \end{enumerate}

 \end{enumerate}

%
%




%

%

\onecolumn
\aistatstitle{Supplementary Material for Mixed-Output Gaussian Process Latent Variable Models}

\appendix
\section{Table of Notation}
\label{appendix: notation}
\begin{table}[ht!]
\begin{center}
\begin{small}
\begin{tabular}{p{0.15\linewidth}  p{0.80\linewidth}}
\toprule
Notation & Definition \\
\midrule
$ \cdot^*$ & Star indicates test samples of a variable (where components are unobserved)  \\
& \textit{Note: there is an exception for $\mu^r_{i\cdot}$, $\Sigma^r_{i\cdot}$ and $\alpha^r_i$ which refer to the test data, but the star is omitted for brevity}\\
$N$ & Number of observations \\
$M$ & Number of observations for each sample \\
$C$ & The number of components that are contained in the samples\\
$L$ & The number of inducing points for each component\\
$A$ & The number of Latent dimensions in the MO-GPLVM model\\
$y_{ij}$ & The observation for sample $i$ at position $j$\\
$r_{ic}$ & The component mixture $c$ for sample $i$ \\ 
$f_{ijc}$ &  The $c$th pure component spectra evaluated at wavelength $j$ for sample $i$\\
$h_i$ & The latent variable of the $i$th sample \\
$\lambda_j$ & The $j$th location of measurement for all the samples\\
$\epsilon_{ij}$ & The Gaussian noise observed for sample $i$ at position $j$\\ 
$r_{i\cdot}$ & The $\mathbb{R}^C$ vector containing all of the mixture components for sample $i$ \\ 
$f_{ij\cdot}$ &  The $\mathbb{R}^C$ vector containing all of the pure component values for sample $i$ at position $j$\\
$Y$ & The $\mathbb{R} ^ {N \times M}$ matrix the training spectra\\
$Y^*$ & The $\mathbb{R} ^ {N^* \times M}$ matrix containing the test spectra\\
$\mathcal{Y}$ & The set of both training and test observations combined \\
$R$ & The $\mathbb{R}^{N\times C}$ matrix containing the training mixture components\\
$R^*$ & The $\mathbb{R}^{N\times C}$ matrix containing the test mixture components\\
$\mathcal{R}$ & The combined training and test weights\\
$F$ & The $\mathbb{R} ^ {N \times M \times C}$ tensor containing all of the training pure component spectra \\
$F^*$ & The $\mathbb{R} ^ {N \times M \times C}$ tensor containing all of the test pure component spectra \\
$\Sc$ & The $\mathbb{R} ^ {N \times M \times C}$ The combined training and test pure component signals \\
$\hat F$ & The $\mathbb{R} ^ {M \times C}$ matrix containing  least squares estimate for the pure component signals are static based on the training set \\
$E$ & The $\mathbb{R} ^ {N \times M}$ matrix containing  the reconstruction error of the training spectra $D$ \\& using static pure components $\hat F$\\
$U$ & The $\mathbb{R}^{L\times C}$ matrix containing of all the inducing points\\
$V$ & The $\mathbb{R}^{L\times (A+1) \times C}$ tensor containing of all the input locations of the inducing points\\
$V_c$ & The $\mathbb{R}^{L\times (A+1)}$  containing the input locations of the inducing points for component $c$\\
$v_{lc}$ &  The input of a specific inducing point\\
$v^{x}_{lc}$ & 
The latent dimensions of inducing point $v_{lc}$\\
$v^{\lambda}_{lc}$ &  The observed dimensions of inducing point $v_{lc}$\\
$u$ & The $\mathbb{R}^{LC}$ vectorized form of $U$\\
$u_c$ & The $\mathbb{R}^{L}$ vector of the output of the inducing points for $c$\\
$\alpha_{i\cdot}$ &  The $\R^c$ vector of the variational parameters for the Dirichlet distribution of $r_{i\cdot}$\\
$y'_{i\cdot}$ & The $\mathbb{R}^{M}$ vector of raw (unpreprocessed) observations of $y_i$\\
$\elbo$ & Evidence Lower Bound (ELBO) of the MO-GPLVM model \\ 
$\mathcal{G}$ & The $U$ dependent terms of $\elbo$\\


\bottomrule
\end{tabular}
\end{small}
\end{center}
\vskip -0.1in
\end{table}
\begin{table}[ht!]
\begin{center}
\begin{small}
\begin{tabular}{rl}
\toprule
Notation & Definition \\
\midrule
$D_{KL}\left(q(\cdot) || p(\cdot)\right) $ & The KL divergence between the distribution $q(\cdot)$ and $p(\cdot)$\\
$\langle g(\cdot)\rangle_{q(\cdot)}$ & The expectation of expression $g(\cdot)$ over the distribution $ q(\cdot) $ \\
$ \mu^h_i$ & Mean of the variational posterior for the position of the latent space\\ & for data point $i$\\
$ \Sigma^h_i$ & The covariance of the variational posterior for the position of the latent space \\ & for data point $i$\\
$\beta_c$ &  Length-scales of the ARD kernel for the latent space for component $c$\\ 
$\gamma_c$ &  Length-scales of the  kernel for the observed space for component $c$\\ 

$\sigma^2_s$ &  Scale of the kernel\\ 
$\sigma^2$ &  Variance of the noise of the Gaussian Process\\ 
$\mu^r_{i\cdot}$ & The $\mathbb{R} ^ C$ mean estimate of component $i$\\
$\Sigma^r_{i\cdot}$ & The $\mathbb{R} ^ {C\times C}$ covariance fore the estimate of $r_{i\cdot}$ \\
$\mu^f_{ij\cdot}$ & The $\mathbb{R} ^ C$ GP mean of $f_{ij\cdot}$ conditional on $U$\\
$K^f_{ij\cdot}$ & The $\mathbb{R} ^ {C\times C}$ covariance fore the estimate of $f_{ij\cdot}$  conditional on $U$\\
$K^f_{\cdot \cdot c}$ & The $\mathbb{R} ^ {NM\times NM}$ covariance for the estimate of $f_{\cdot \cdot c}$  conditional on $u_c$\\
$K^c_{\bullet \bullet }$ & The $\mathbb{R} ^ {NM\times NM}$ covariance where each element is given by $k^c((h_i, \lambda_j),(h_{i'}, \lambda_{j'}))$\\
$K^c_{\bullet V_c }$ & The $\mathbb{R} ^ {NM\times L}$ covariance where each element is given by $k^c((h_i, \lambda_j),v_{lc})$\\
$K^c_{V_c V_c }$ & The $\mathbb{R} ^ {L\times L}$ covariance where each element is given by $k^c(v_{lc},v_{lc})$\\
$K_{VV}$ & The $\mathbb{R}^{LC \times LC }$ kernel matrix giving the covariance of the inducing points. \\ & \textit{Note: This has block diagonal structure of $L\times L$ matrices }\\
$K_{(h_i, \lambda_j)V}$ & The $\mathbb{R}^{C \times LC }$ kernel matrix giving the covariance between $f_{ij\cdot}$ and the inducing points. \\ 
& \textit{Note: This has block diagonal structure of $1\times L$ matrices.}\\
$K_{(h_i, \lambda_j)(h_i \lambda_j)}$ & The $\mathbb{R}^{C \times C }$ kernel matrix giving the covariance between $f_{ij\cdot}$ and the inducing points. \\ 
& \textit{Note: This has a diagonal structure.}\\
$\xi_0$ & See Equation~\eqref{xi_0}. Analogous to $\psi_0$ in the regular GPLVM model.\\ 
$\xi_1$ & See Equation~\eqref{xi_1}. Analogous to $\psi_1$ in the regular GPLVM model.\\ 
$\xi_2$ & See Equation~\eqref{xi_2}. Analogous to $\psi_2$ in the regular GPLVM model.\\ 
$\delta (\cdot)$ & The Dirac delta function.\\
\bottomrule
\end{tabular}
\end{small}
\end{center}
\vskip -0.1in
\end{table}
\newpage
\section{ ELBO Derivations} 
\subsection{Full ELBO derivation} \label{appendix: ELBO derivation}
We start from the ELBO for our variational distribution:
\begin{equation}
    \begin{split}
        \elbo = &  \int \log\left(\frac{p(  Y|  R,   F) p(  R)   p( Y^*|R^*, F^*) p(F, F^*| H, H^*, U)  p(  H) p(H^*) p(U) }{  q(F,F^*| H,  H^*, U)q(R^*)  q(H) q(H^*) q(U)}\right)\\&  q(F,F^*| H^*, H^*, U) q(H) q(H^*) q(R^*) q(U) dR^*  dF dF^* dH  dH^* dU. 
    \end{split}
\end{equation}
By setting $q(F ,F^*| H, U) = p(F, F^*| H, H^*, U)$ the above simplifies to 
\begin{equation}
    \begin{split}
        \elbo = &  \int \log\left(\frac{p(  Y|  R,   F)p(  R)  p(  H) p( Y^*|R^*, F^*)p(R^*)  p(H^*) p(U) }{  q(R^*) q(H^*)  q(H) q(U)}\right)  p(F, F^*| H, H^*, U) q(H) q(H^*) \\ & \; q(R^*)   
 q(U) dR^* dF dF^* dH dH^*  dU 
        \\= &  \underbrace{\int \log\left(\frac{p(  Y|  R,   F)   p( Y^*|R^*, F^*) p(U)  }{    q(U)}\right) p(F, F^*| H, H^*, U) q(H) q(H^*) q(R^*) q(U) dU dF dF^* dH dH^* dR^*}_{\mathcal{G}}
        \\ & - D_{KL} \left(q(H)|| p(H)\right) - D_{KL} \left(q(H^*)|| p(H^*)\right) - D_{KL} \left(q(R^*)|| p(R^*)\right)  + \log(p(  R)).
    \end{split}
\end{equation}
The challenging terms to deal with are those that depend on $U$, we therefore collect these into the new term $\mathcal{G}$ which we now focus on. 

\begin{equation}
    \begin{split} 
        \mathcal{G}   = &    \int \frac{-1}{2\sigma^2}\bigg( \sum_{  i =1}^{  N} \sum_{j=1}^M ( y_{  i j} - r^T_{  i \cdot} f_{  i j \cdot})^2 + \sum_{i =1}^{ N^*} \sum_{j=1}^M ( y^*_{ i j} - r^{* \; T}_{ i \cdot} f^{*}_{ i j \cdot})^2  \bigg) 
        \\ & \quad p(F, F^*| H, H^*, U) q(H) q(H^*) q(R^*) q(U) dU dF dF^* dH dH^* dR^*   
        \\&  \qquad -  \frac{NM}{2} \log(\sigma^2) - \frac{NM}{2}log(2\pi) - D_{KL}\left(q(U)|| p(U)\right)
    \end{split}
\end{equation}

We next take the expectation over $R^*$, for which we introduce
\begin{equation}
    \langle r^*_{i \cdot} \rangle_{q(R^*)} = \mu^r_{i\cdot}
\end{equation}
and 
\begin{equation}
\begin{split}
    \langle r^*_{i \cdot} r^{* \; T}_{i \cdot}  \rangle_{q(r^*)} = & \langle ((r^*_{i \cdot} - \mu^r_{i\cdot}) + \mu^r_{i\cdot})  (r^*_{i \cdot} - \mu^r_{i\cdot} )+ \mu^{r \; T}_{i\cdot})) \rangle_{q(r^*)} 
    \\ = & \langle (r^*_{i \cdot} - \mu^r_{i\cdot})(r^*_{i \cdot} - \mu^r_{i\cdot} )^T  + 2 \cancel{\mu^r_{i\cdot}  (r_{i \cdot} - \mu^r_{i\cdot} )} + {\mu^r_{i\cdot}\mu^r_{i\cdot}}^T  \rangle_{q(r^*)} 
    \\ = & \Sigma^r_{i\cdot}  + {\mu^r_{i\cdot} \mu^r_{i\cdot}}^T  
\end{split}
\end{equation}
for the Dirichlet distribution $Dir(r_{i \cdot}|\alpha_{i\cdot})$, which is our primary focus,  these quantities are given by 
\begin{equation}
    \mu^r_{i\cdot} = (\mu^r_{i1}, \dots, \mu^r_{iC})^T,  \quad \text{where} \quad \mu^r_{ic} = \frac{\alpha_{ic}}{\sum_{c=1}^C \alpha_{ic}}
\end{equation}
\begin{equation}
    \Sigma^r_{i\cdot} = \frac{1}{1 + \sum_{c=1}^C \alpha_{ic} }\left(\text{diag}(\mu^r_{i\cdot}) -\mu^r_{i\cdot }\mu^{r\; T}_{i\cdot }\right).
\end{equation}

For the mixture model the classification distribution is the multinomial distribution with one observation, so
\begin{equation}
    \mu^r_{i\cdot} = (\alpha_1, \dots, \alpha_C), \quad \text{where} \quad \sum_{c=1}^C \alpha_{ic} = 1 \; \forall  \;i
\end{equation}
\begin{equation}
    \Sigma^r_{i\cdot} = \text{diag}(\mu^r_{i\cdot}) -\mu^r_{i\cdot }\mu^{r\; T}_{i\cdot }
\end{equation}

These allow us to write 
\begin{equation}
    \begin{split} 
        \mathcal{G}  & =   \int \frac{-1}{2\sigma^2}\bigg( \sum_{  i =1}^{  N} \sum_{j=1}^M  y_{  i j}^2  - 2 y_{  i j}r^T_{  i \cdot} f_{  i j \cdot}  + f_{  i j \cdot}^T r_{  i \cdot} r_{  i \cdot}^T f_{  i j \cdot}
        \\ & \qquad  +  \sum_{ i =1}^{N^*} \sum_{j=1}^M  {y^*_{i j}}^{ \; 2}  - 2y^*_{ij}\mu^{r\;T}_{ i \cdot} f^*_{ i j \cdot} + 
        Tr\left(\left(\mu^r_{ i \cdot}\mu^{r \; T}_{ i \cdot} + \Sigma^r_{i\cdot}\right)f^*_{i j \cdot} f_{ i j \cdot}^{*\; T}\right)\bigg)
        p(F, F^*| H, H^*, U)  \\ & \qquad q(H) q(H^*) q(U)  dF dF^* dH dH^*   dU  
        \\& \qquad -  \frac{NM}{2} \log(\sigma^2) - \frac{NM}{2}log(2\pi) - D_{KL}\left(q(U)|| p(U)\right).
    \end{split}
\end{equation}

The next step is to take the expectation over $F$. At this point it is should be noted that, while the prior factorizes over each component $c$, the likelihood factorizes over each observation $ij$ which is the combination of  $C$ Gaussian Processes. For this derivation we have been working with the likelihood, it is therefore necessary to be able to write the vector $f_{ij\cdot}$ as a linear transformation of of $U$ which requires some new quantities that we introduce now. Firstly, we introduce $u\in \R^{LC} = \text{vec}(U)$ which is the vectorized form of the inducing point outputs $U$. To go with this we use the block diagonal matrices
\begin{equation} \label{eq: K_VV}
    K_{VV}\in R^{LC\times LC} = \begin{pmatrix}
        K^1_{V_1 V_1} & \dots &  0
        \\ \vdots & \ddots & \vdots
        \\ 0 & \dots & K^C_{V_C V_C} 
    \end{pmatrix} \quad \text{with elements }  \quad [K^c_{V_c V_c}]_{l l'} = k_c(v_{lc}, v_{l'c})
\end{equation} 
\begin{equation} \label{eq: K_xi lambda j V}
    K_{(h_i, \lambda_j)V}\in R^{C\times LC} = \begin{pmatrix}
        K^1_{(h_i, \lambda_j) V_1} & \dots &  0
        \\ \vdots & \ddots & \vdots
        \\ 0 & \dots & K^C_{(h_i, \lambda_j) V_C}
    \end{pmatrix} \; \text{with elements}  \; [K^c_{(h_i, \lambda_j) V_c}]_{1 l} = k_c((h_i, \lambda_j), v_{lc}), 
\end{equation} 
and 
\begin{equation} \label{eq: K_xi lambdaj xi lambdaj}
    \begin{split} 
        K_{(h_i, \lambda_j)(h_i, \lambda_j)}\in R^{C\times C} = \begin{pmatrix}
        k_1((h_i, \lambda_j), (h_i, \lambda_j)) & \dots &  0
        \\ \vdots & \ddots & \vdots
        \\ 0 & \dots & k_C((h_i, \lambda_j), (h_i, \lambda_j))
    \end{pmatrix}.
    \end{split}
\end{equation}
These matrices allow us to write expressions for the expectations over $F$ as  
\begin{equation}
    \langle f_{ij\cdot} \rangle_{p(F|U,H)} = \mu^f_{ij\cdot} = K_{(h_i, \lambda_j) V} K_{VV}^{-1} u
\end{equation}
\begin{equation}
        \langle f_{ij\cdot} f_{ij\cdot}^T \rangle_{p(F|U,H)} = K^f_{ij\cdot} + \mu^f_{ij\cdot}\mu^{f\; T}_{ij\cdot} \quad \text{where} \quad K^f_{ij\cdot} = K_{(h_i, \lambda_j)(h_i, \lambda_j)} - K_{(h_i, \lambda_j)V} K_{VV}^{-1} K_{(h_i, \lambda_j)V}^T.
\end{equation}

Using these results we continue the derivation as 
\begin{equation}
    \begin{split} 
        \mathcal{G}  & =   \int \frac{-1}{2\sigma^2}\bigg( \sum_{  i =1}^{  N} \sum_{j=1}^M  - 2 y_{  i j}r^T_{  i \cdot} \mu^f_{  i j \cdot}  +  Tr\left(r_{  i \cdot} r_{  i \cdot}^T \left(\mu^f_{  i j \cdot} \mu^{f \; T}_{  i j \cdot} + K^f_{ij\cdot}\right)\right)
        \\ & + \sum_{ i =1}^{N^*} \sum_{j=1}^M  - 2y_{ij}\mu^{r\;T}_{ i \cdot} \mu ^{ f^*}_{ i j \cdot} +  Tr\left(\left(\mu^r_{i \cdot}\mu^{r \; T}_{ i \cdot} + \Sigma^r_{i\cdot}\right)\left(\mu^{f^*}_{ i j \cdot} \mu^{f^*\; T }_{ i j \cdot} + K^{f^*}_{ij\cdot}\right)\right)\bigg)
        q(H) q(H^*) q(U)  dH dH^*  dU  
        \\ & - \frac{1}{2\sigma^2 }\left(\sum_{i=1}^N \sum_{j=1}^M y_{i j}^2 + \sum_{i=1}^{N^*} \sum_{j=1}^M y^{*\; 2}_{i j}\right) -  \frac{NM}{2} \log(\sigma^2) - \frac{NM}{2}log(2\pi) - D_{KL}\left(q(U)|| p(U)\right)
        \\ & =   \int \frac{-1}{2\sigma^2}\bigg( \sum_{  i =1}^{  N} \sum_{j=1}^M    - 2 y_{  i j}r^T_{  i \cdot} K_{(h_{  i} \lambda_j)V }K_{VV}^{-1} u 
   + u^T K^{-1}_{VV} K^T_{(h_i, \lambda_j)V} r_{  i \cdot} r_{  i \cdot}^T K_{(h_i, \lambda_j)V }K_{VV}^{-1} u  
   \\ & + Tr\left( r_{  i \cdot} r_{  i \cdot}^T 
   \left(K_{(h_i, \lambda_j) (h_i, \lambda_j)} - K_{(h_i, \lambda_j) V} K^{-1}_{VV} K^T_{(h_i, \lambda_j) V}\right)\right)
        \\ & + \sum_{ i =1}^{ N^*} \sum_{j=1}^M  - 2y_{ij}\mu^{r\;T}_{ i \cdot} K_{(h^*_i \lambda_j)V }K_{VV}^{-1} u + 
 u^T K^{-1}_{VV} K^T_{(h^*_i\lambda_j)V} \left(\mu^r_{ i \cdot}\mu^{r \; T}_{ i \cdot} + \Sigma^r_{i\cdot}\right)K_{(h^*_i\lambda_j)V }K_{VV}^{-1} u   
 \\ & + Tr\left(\left(\mu^r_{ i \cdot}\mu^{r \; T}_{ i \cdot} + \Sigma^r_{i\cdot}\right)\left(K_{(h^*_i, \lambda_j) (h^*_i, \lambda_j)} - K_{(h^*_i, \lambda_j) V} K^{-1}_{VV} K^T_{(h^*_i, \lambda_j) V}\right)\right)\bigg)
        \\ & \qquad   q(H)q(H^*) q(U) dH dH^*  dU   - \frac{1}{2\sigma^2 }\left(\sum_{i=1}^N \sum_{j=1}^M y_{i j}^2 + \sum_{i=1}^{N^*} \sum_{j=1}^M y^{*\; 2}_{i j}\right)\\& -  \frac{NM}{2} \log(\sigma^2) - \frac{NM}{2}log(2\pi) - D_{KL}\left(q(U)|| p(U)\right).
    \end{split}
\end{equation}

By performing some algebraic manipulations it is possible to arrive at the form
\begin{equation} \label{eq: deviation between derivations}
    \begin{split} 
        \mathcal{G}   = &   \int \frac{-1}{2\sigma^2}\Biggl(  - 2\left( \sum_{  i =1}^{  N} \sum_{j=1}^M   y_{  i j}r^T_{  i \cdot} K_{(h_i, \lambda_j)V} + \sum_{  i =1}^{  N^*} \sum_{j=1}^M  y^*_{ij}\mu^{r\;T}_{  i \cdot} K_{(h^*_i, \lambda_j)V }\right)K_{VV}^{-1} u 
        \\ & + u^T K^{-1}_{VV} \left(\sum_{i=1}^N \sum_{j=1}^M K^T_{(h_i, \lambda_j)V} r_{  i \cdot} r_{  i \cdot}^T K_{(h_i, \lambda_j)V } + \sum_{i=1}^{N^*} \sum_{j=1}^M K^T _{(h^*_i, \lambda_j)V} \left(\mu^r_{  i \cdot}\mu^{r \; T}_{  i \cdot} + \Sigma^r_{i\cdot}\right) K _{(h^*_i, \lambda_j) V} \right)K_{VV}^{-1} u 
        \\ & + \sum_{  i =1 }^{  N}\sum_{j =1 }^{M} Tr\left( r_{  i \cdot} r_{  i \cdot}^T \left( K_{(h_{  i}, \lambda_j)(h_{  i}, \lambda_j)}   - K_{(h_{  i} \lambda_j) V }^Tr_{  i \cdot} r_{  i \cdot}^TK_{(h_{  i} \lambda_j) V } K_{VV}^{-1} \right) \right)
       \\ & + \sum_{ i =1 }^{ N^*}\sum_{j =1 }^{M} Tr\left(\left(\mu^r_{i \cdot}\mu^{r \; T}_{i \cdot} + \Sigma^r_{i\cdot}\right)\left( K_{(h^*_{i} \lambda_j)(h^*_{ i} \lambda_j)}  - K_{(h_{i} \lambda_j) V } K_{VV}^{-1} K_{(h_{i} \lambda_j) V }^T\right)\right)\Biggr)  q(H) q(H^*) q(U) dH dH^* dU 
       \\ & - \frac{1}{2\sigma^2 }\left(\sum_{i=1}^N \sum_{j=1}^M y_{i j}^2 + \sum_{i=1}^{N^*} \sum_{j=1}^M y^{*\; 2}_{i j}\right) -  \frac{NM}{2} \log(\sigma^2) - \frac{NM}{2}log(2\pi) - D_{KL}\left(q(U)|| p(U)\right).
 	\end{split}
\end{equation}
which gathers the terms for which we need to compute the expectation over $H$ and $H^*$.
We can write these terms using three quantities which we introduce here: 
\begin{equation}
\begin{split} \label{xi_0}
    \xi_0 = & \left \langle \sum_{  i =1 }^{  N}\sum_{j =1 }^{M} Tr\left( r_{  i \cdot} r_{  i \cdot}^T K_{(h_{  i} \lambda_j )(h_{  i} \lambda_j )}\right) + \sum_{ i =1 }^{ N^*}\sum_{j =1 }^{M} \left(Tr\left(\left(\mu^r_{ i \cdot}\mu^{r \; T}_{ i \cdot} + \Sigma^r_{i \cdot }\right)K_{(h^*_{  i} \lambda_j )(h^*_{  i} \lambda_j )}\right)\right)\right \rangle_{q(  H, H^*)}
\end{split}
\end{equation}

\begin{equation} \label{xi_1}
\begin{split}
    \xi_1  =& \left \langle \sum_{  i =1}^{  N} \sum_{j=1}^M   y_{  i j}r^T_{  i \cdot} K_{(  h_i \lambda_j)V} + \sum_{ i =1}^{ N^*} \sum_{j=1}^M  y^*_{ij}\mu^{r\;T}_{  i \cdot} K_{( h^*_i \lambda_j)V } \right \rangle_{q(  H, H^*)} \\
    = & \sum_{  i =1}^{  N} \sum_{j=1}^M  y_{  i j}r^T_{  i \cdot} \left \langle K_{(h_i, \lambda_j)V} \right \rangle_{q(  H)}+ \sum_{ i =1}^{ N^*} \sum_{j=1}^M  y^*_{ij}\mu^{r\;T}_{ i \cdot} \left \langle K_{(h^*_i \lambda_j)V } \right \rangle_{q(H^*)} \\
\end{split}
\end{equation}
\begin{equation} 
\begin{split}\label{xi_2}
    \xi_2 = &   \left \langle  \sum_{  i = 1 }^{  N} \sum_{j=1}^M K_{(h_{  i} \lambda_j)V}^Tr_{  i \cdot} r_{  i \cdot}^T K_{( h_{  i} \lambda_ j)V } + \sum_{i = 1 }^{N^*} \sum_{j=1}^M K^T _{(h^*_{i} \lambda_j)V} \left(\mu^r_{  i \cdot}\mu^{r \; T}_{  i \cdot} + \Sigma^r_{i\cdot}\right) K _{(h_i \lambda_j)V} \right\rangle_{q(  H, H^*)} 
    \\ = &  \sum_{  i = 1 }^{  N} \sum_{j=1}^M \left \langle K^T_{(h_{  i} \lambda_j)V } r_{  i \cdot} r_{  i \cdot}^T K_{(h_i \lambda_j)V } \right \rangle_{q(  H)}  + \sum_{i=1}^{N^*} \sum_{j=1}^M \left \langle K^T_{(h^*_{ i} \lambda_j )V} \left(\mu^r_{ i \cdot}\mu^{r \; T}_{ i \cdot} + \Sigma^r_{i\cdot}\right) K_{(h^*_{i} \lambda _j )V} \right \rangle_{q(H^*)} 
\end{split}
\end{equation}

Using these quantities and the the fact that $p(u) = N(u| 0, K_{VV})$ we can write
\begin{equation}
    \begin{split} 
        \mathcal{G}  =&   \int \bigg(  \frac{1}{\sigma^2}\xi_1 K_{VV}^{-1} u -\frac{1}{2} u^TK^{-1}_{VV}\left( \frac{-1}{\sigma^2} \xi_2 +   K_{VV} \right)K_{VV}^{-1} u \bigg)  \log\left(\frac{1}{q(u)} \right)q(U) \, dU- \frac{M}{2}log(2 \pi) 
        \\ & -  \frac{1}{2}log(det(K_{VV})) + \frac{1}{2\sigma^2}  Tr\left(K^{-1}_{VV}\xi_{2}\right) + \frac{1}{2 \sigma^2}\xi_0- \frac{1}{2\sigma^2 }\left(\sum_{i=1}^N \sum_{j=1}^M y_{i j}^2 + \sum_{i=1}^{N^*} \sum_{j=1}^M y^{*\; 2}_{i j}\right) 
        \\ & -  \frac{NM}{2} \log(\sigma^2) - \frac{NM}{2}log(2\pi).
    \end{split}
\end{equation}

We can use the approach from \citet{king2006fast} to find the optimal distribution for $u$.  We do this by noting  that the expression in the integral can be rewritten as a quadratic form with the quantities  $\Sigma_u = K_{VV}\left( \frac{1}{\sigma^2} \xi_2 +   K_{VV} \right)^{-1}K_{VV}$ and $\mu_u = \Sigma_u K^{-1}_{VV} \xi_1$, giving
\begin{equation}
    \begin{split} 
        \mathcal{G}   = & \int - \frac{1}{2} (u - \mu_u )^T\Sigma_u^{-1}(u - \mu_u ) \log\left(\frac{1}{q(u)}\right) q(u)du + \frac{M}{2}log(2 \pi) -  \frac{1}{2}log(det(K_{VV}))
        \\  & +   \frac{1}{2\sigma^4} \xi_1^T  \left(\frac{1}{\sigma^2} \xi_2 + K_{VV}\right)^{-1}  \xi_1   - \frac{1}{2}log\left(det\left(\frac{1}{\sigma^2} \xi_2 + K_{VV}\right)\right)  + \frac{1}{2} \log( det(K_{VV})) 
        \\ &  - \frac{1}{2 \sigma^2}  \xi_{0} + \frac{1}{2\sigma^2}  Tr\left(K^{-1}_{VV}\xi_{2}\right) - \frac{1}{2\sigma^2 }\left(\sum_{i=1}^N \sum_{j=1}^M y_{i j}^2 + \sum_{i=1}^{N^*} \sum_{j=1}^M y^{*\; 2}_{i j}\right)   -  \frac{NM}{2} \log(\sigma^2) - \frac{NM}{2}log(2\pi).
    \end{split}
\end{equation}

 By setting $q(u) \sim N(\mu^u, \Sigma^u)$ and performing some algebra we arrive at the final form of $\mathcal{G}$ of 
\begin{equation}
    \begin{split} 
        \mathcal{G}  & =   \frac{1}{2\sigma^4} \xi_1^T  \left(\frac{1}{\sigma^2} \xi_2 + K_{VV}\right)^{-1}  \xi_1   - \frac{1}{2}log\left(det\left(\frac{1}{\sigma^2} \xi_2 + K_{VV}\right)\right)  + \frac{1}{2} \log( det(K_{VV})) 
        \\ &  - \frac{1}{2 \sigma^2}  \xi_{0} + \frac{1}{2\sigma^2}  Tr\left(K^{-1}_{VV}\xi_{2}\right) - \frac{1}{2\sigma^2 }\left(\sum_{i=1}^N \sum_{j=1}^M y_{i j}^2 + \sum_{i=1}^{N^*} \sum_{j=1}^M y^{*\; 2}_{i j}\right)   -  \frac{NM}{2} \log(\sigma^2) - \frac{NM}{2}log(2\pi). 
    \end{split}
\end{equation}
By combining this with the terms which are independent of $u$ that we set aside initially we arrive at 
\begin{equation}
    \begin{split} 
        \mathcal{L}   = &  \frac{1}{2\sigma^4} \xi_1^T  \left(\frac{1}{\sigma^2} \xi_2 + K_{VV}\right)^{-1}  \xi_1   - \frac{1}{2}log\left(det\left(\frac{1}{\sigma^2} \xi_2 + K_{VV}\right)\right)  
         - \frac{1}{2 \sigma^2}  \xi_{0}
        \\ &+ \frac{1}{2\sigma^2}   Tr\left(K^{-1}_{VV}\xi_{2}\right)  - \frac{1}{2\sigma^2 }\left(\sum_{i=1}^N \sum_{j=1}^M y_{i j}^2 + \sum_{i=1}^{N^*} \sum_{j=1}^M y^{*\; 2}_{i j}\right)   -  \frac{NM}{2} \log(\sigma^2) 
        - \frac{NM}{2}log(2\pi) 
        \\ & + \frac{1}{2} \log( det(K_{VV}))  - KL\left(q(h)||p(h)\right) - KL\left(q(R^*)||p(R^*)\right).
    \end{split}
\end{equation}

\subsection{Derivation of the ELBO with Independent Measurements } \label{appendix: independent measurements ELBO}


\textit{Note:} The notation for this derivation differs slightly from the derivation above and the table given in Appendix~\ref{appendix: notation}. Where there are differences they are stated here.

In this derivation we present a derivation of the ELBO where we do not assume that the measurement is smooth and can be described by a Gaussian Process across the observed inputs. The change in the Gaussian Process prior is that it it now factorizes across the $MC$ vectors of $f_{\cdot j c } \in \R^N$ 
\begin{equation}
    p(F) = \prod_{c=1}^C\prod_{j=1}^M N(f_{\cdot j c}| 0, K^{jc}_{V_{jc}V_{jc}}). 
\end{equation}

As we are working with the variational sparse changing the prior for $F$ requires changing the inducing points and their prior such that they are drawn from the output of the GP.  This requires increasing the number of inducing points from $LC$ to $MLC$, as now the Gaussian process factorizes over the $M$ observations. 
To account for this we change the notation in this section so that  $U\in \R^{M\times L\times C}$, made up of $MC$ sets of $u_{jc} \in \R^L$, and $V \in \R^{M\times L \times C \times A}$, made up of $MC$ sets of $u_{jc} \in \R^{L\times A}$, and has a prior
\begin{equation}
    p(U) = \prod_{c=1}^C\prod_{j=1}^M  N\left(u_{jc}|0, K^{jc}_{V_{jc}V_{jc}}\right), 
\end{equation}
where the kernel matrix is now a $N\times N$ matrix with elements 
\begin{equation}
   \left[K^{jc}_{V_{jc}V_{jc}}\right]_{i i'} = k_{jc}(h_i, h_{i'}).
\end{equation}

As it is impractical to find independent kernels and inducing points for each of the measurements and locations, for the remainder of this derivation we assume that $k_{jc}(h_i, h_{i'}) = k_c(h_i, h_{i'})$ and $V_{jc} = V$ for all $j$ and $c$, allowing us to maintain the same definitions for the kernel matrix $K_{VV}$ as defined in Equation~\eqref{eq: K_VV}, and the kernel matrices defined in Equation~\eqref{eq: K_xi lambda j V} and Equation~\eqref{eq: K_xi lambdaj xi lambdaj} only need to be adapted to $K_{h_i V}$ and $K_{h_i h_i}$ by changing the kernel function to 
\begin{equation}
    k_c(h_i, h_{i'}) = \sigma^2_{f_c} exp\left(\frac{1}{2} (h_i - h_{i'})^T \beta^{-1}(h_i - h_{i'})\right).
\end{equation}

With this the derivations are identical until Equation~\eqref{eq: deviation between derivations}, which is changed to include the fact that now each of the measurement locations depends on a different set of inducing points $u_j \in \R^{LC}$. This modifies the equation to be 
\begin{equation} 
    \begin{split} 
        \mathcal{G}   = &  \sum_{j=1}^M \Biggl[\int \frac{-1}{2\sigma^2}\Biggl(  - 2  \left( \sum_{  i =1}^{  N}    y_{  i j}r^T_{  i \cdot} K_{h_iV} + \sum_{  i =1}^{  N}  y_{ij}\mu^{r\;T}_{  i \cdot} K_{h_iV }\right)K_{VV}^{-1} u_{j}
        \\ & +  u^T_{j} K^{-1}_{VV} \left(\sum_{i=1}^N  K^T_{h_iV} r_{  i \cdot} r_{  i \cdot}^T K_{h_iV } + \sum_{i=1}^{N^*} K^T _{h_i V} \left(\mu^r_{  i \cdot}\mu^{r \; T}_{  i \cdot} + \Sigma^r_{i\cdot}\right) K _{h_i V} \right)K_{VV}^{-1} u_j 
        \\ & + \sum_{  i =1 }^{  N}\sum_{j =1 }^{M} Tr\left( r_{  i \cdot} r_{  i \cdot}^T  K_{h_{  i}h_{  i}} \right) - Tr\left(K_{h_{  i}  V }^Tr_{  i \cdot} r_{  i \cdot}^TK_{h_{  i}  V } K_{VV}^{-1} \right) 
       \\ & + \sum_{ i =1 }^{ N^*}\sum_{j =1 }^{M} Tr\left(\left(\mu^r_{i \cdot}\mu^{r \; T}_{i \cdot} + \Sigma^r_{i\cdot}\right) K_{h_{i} h_{ i}} - Tr\left(K_{h_{i}  V } K_{VV}^{-1} K_{h_{i}V }^T\right)\right)\Biggr)  q(H) q(H^*) q(U) dH dU 
       \Biggr] 
       \\ & - \frac{1}{2\sigma^2 }\left(\sum_{i=1}^N \sum_{j=1}^M y_{i j}^2 + \sum_{i=1}^{N^*} \sum_{j=1}^M y^{*\; 2}_{i j}\right) -  \frac{NM}{2} \log(\sigma^2) - \frac{NM}{2}log(2\pi) - D_{KL}\left(q(U)|| p(U)\right) .
 	\end{split}
\end{equation}

Note that the terms inside of the sum are identical to those those in the correlated measurements case, once the kernel matrices and inducing points have been redefined. So, by simply redefining the values of $\xi_i$ to remove dependence on $\lambda$ all of the same arguments as above hold. These new terms, equivalent to Equations~\eqref{xi_0}-\eqref{xi_2}, are
\begin{equation}
\begin{split} 
    \xi_0 = &  \sum_{  i =1 }^{  N} \left \langle Tr\left( r_{  i \cdot} r_{  i \cdot}^T K_{(h_{  i} \lambda_j )(h_{  i} \lambda_j )}\right) \right \rangle_{q(H)} 
     + \sum_{ i =1 }^{ N^*}\left \langle Tr\left(\left(\mu^r_{ i \cdot}\mu^{r \; T}_{ i \cdot} + \Sigma^r_{i \cdot }\right)K_{h_{  i}h_{  i} }\right)\right \rangle_{q(H^*)} \in \R 
\end{split}
\end{equation}

\begin{equation}
\begin{split} 
    \xi_{1j}  
    = & \sum_{  i =1}^{  N} y_{  i j}r^T_{  i \cdot} \left \langle K_{h_i V} \right \rangle_{q(  H)}+ \sum_{ i =1}^{ N^*}  y^*_{ij}\mu^{r\;T}_{ i \cdot} \left \langle K_{h_i V } \right \rangle_{q(H^*)} \in \R^{C \times MC } \\
\end{split}
\end{equation}
and
\begin{equation} 
\begin{split}
    \xi_2 = &  \sum_{  i = 1 }^{  N}  \left \langle K^T_{h_{  i} V } r_{  i \cdot} r_{  i \cdot}^T K_{h_i V } \right \rangle_{q(  H)}  + \sum_{i=1}^{N^*} \left \langle K^T_{h_{ i} V} \left(\mu^r_{ i \cdot}\mu^{r \; T}_{ i \cdot} + \Sigma^r_{i\cdot}\right) K_{h_{i} V} \right \rangle_{q(H^*)} . \in \R ^{MC \times MC}
\end{split}
\end{equation}
Note that (provided the same kernel and inducing points are used for all $j$) the expressions for $\xi_0$ and $\xi_2$ are both independent of the measurement location, whereas $\xi_{1j}$ does depend on the values measured at the specific measurement locations - demanding the additional index $j$. 

Using this insight we skip to the final definition of the ELBO as

\begin{equation}
    \begin{split} 
        \mathcal{L}   = & \sum_{j=1}^M \Biggl[\frac{1}{2\sigma^4} \xi_{1j}^T  \left(\frac{1}{\sigma^2} \xi_2 + K_{VV}\right)^{-1}  \xi_{1j}   - \frac{1}{2}log\left(det\left(\frac{1}{\sigma^2} \xi_2 + K_{VV}\right)\right)  
         - \frac{1}{2 \sigma^2}  \xi_{0} + \frac{1}{2\sigma^2}   Tr\left(K^{-1}_{VV}\xi_{2}\right) \Biggr]
        \\ & - \frac{1}{2\sigma^2 }\left(\sum_{i=1}^N \sum_{j=1}^M y_{i j}^2 + \sum_{i=1}^{N^*} \sum_{j=1}^M y^{*\; 2}_{i j}\right)   -  \frac{NM}{2} \log(\sigma^2) 
        - \frac{NM}{2}log(2\pi) 
         + \frac{1}{2} \log( det(K_{VV}))  
         \\&- KL\left(q(H)||p(H)\right) - KL\left(q(H^*)||p(H^*)\right) - KL\left(q(R^*)||p(R^*)\right).
    \end{split}
\end{equation}

\subsubsection{Relative Computational Cost}

While increasing the number of latent variables from $LC$, as for the first ELBO derivation, to $LMC$, for the derivation presented here, may sound computationally intensive, the new inducing points now only need to describe the variation across latent variables of a single  of a single measurement location, rather than needing to describe the variation across all the latent values and all the measurements together. As the information of a single measurement location is clearly significantly less than that for all the measurements collectively, a drastically lower number of inducing points can be used to describe each of the vectors $f_{\cdot j c } \in \R^N$ than would be required to describe $f_{\cdot \cdot c } \in \R^{NM}$. Furthermore, as working with the MO-GPLVM requires inverting a $LC\times LC$ matrix, it is generally more computationally efficient to work considering all of the $M$ observations to be independent - although at the cost of throwing away the information associated with the measurement locations. 

\section{Expressions for Weighted Kernel Expectations} \label{Appendix: xi}

The $\xi_i$ values are expectations over three kernel matrices, with expressions given in Equation~\eqref{xi_0}-\eqref{xi_2}. For this work we have only considered a single ARD kernel shared between all of the components as this allows us to calculate the expressions for $\xi_i$ analytically, however it is possible to use other Kernels and compute the expectation using numerical methods \citep{de2021learning, lalchand2022generalised}. 

For the ARD kernels $k_c(\cdot, \cdot) = k(\cdot, \cdot ) \forall c = 1, ..., C$ 
\begin{equation}
     k((h_i, \lambda_j),  (h_{i'}, \lambda_{j'})) =
    \sigma^2_{s} exp\left(-\frac{1}{2}\left(h_i - h_{i'}\right)^T \beta^{-1} \left(h_i - h_{i'}\right)\right) exp\left(-\frac{1}{2}\left(\lambda_j - \lambda_{j'}\right)^T \gamma^{-1} \left(\lambda_j - \lambda_{j'}\right)\right) 
\end{equation}
the values are given by 
\begin{equation}
    \xi_0 = M \left( \sum_{c=1}^C \sigma^2_{s}\left(\sum_{i = 1 }^ N  r_{ic}^2   + \sum_{i = 1 }^ {N^*} \mu^{r\;2}_{ic}  + [\Sigma^r_{i\cdot}]_{cc})\right)\right)
\end{equation}
\begin{equation}
\begin{split}
    &\xi_1 = \begin{pmatrix}
        [\xi_{1}]_1 & \dots &  0
        \\ \vdots & \ddots & \vdots
        \\ 0 & \dots & [\xi_{1}]_C
    \end{pmatrix} \; \text{where}  \; [\xi_{1}]_c\in \R^{1\times L}
\end{split}
\end{equation}
and the elements of 
\begin{equation}
\begin{split}
    &[[\xi_{1}]_c]_{1 l} =  \sum_{j = 1}^M \sum _{i=1}^N y_{ij} r_{ic}\sigma^2_{f_c} \sqrt{\frac{\det(\beta)}{\det\left(\beta_c + \Sigma^h_{i}\right)}} exp\left(-\frac{1}{2} (\mu^h_{i} - v^h_{lc})^T \left(\beta + \Sigma^h_{i}\right)^{-1} (\mu^h_{i} - v^h_{lc}) +  (\lambda_{j} - v^{\lambda}_{lc})^T \gamma^{-1} (\lambda_{j} - v^{\lambda}_{lc}) \right)
    \\ & + \sum_{j = 1}^M \sum _{i=1}^{N^*} y^*_{ij} \mu^r_{ic}\sigma^2_{f_c} \sqrt{\frac{\det(\beta)}{\det\left(\beta + \Sigma^{x^*}_{i}\right)}} exp\left(-\frac{1}{2} (\mu^{x^*}_{i} - v^{x^*}_{lc})^T \left(\beta + \Sigma^{x^*}_{i}\right)^{-1} (\mu^{x^*}_{i} - v^{x^*}_{lc}) +  (\lambda_{j} - v^{\lambda}_{lc})^T \gamma^{-1} (\lambda_{j} - v^{\lambda}_{lc}) \right)
\end{split}
\end{equation}
\begin{equation}
\begin{split}
    &\xi_2 = \begin{pmatrix}
        [\xi_{2}]_{11} & \dots &  [\xi_{2}]_{1C}
        \\ \vdots & \ddots & \vdots
        \\ [\xi_{2}]_{C1} & \dots & [\xi_{2}]_{CC} 
    \end{pmatrix} \; \text{where}  \; \xi_{2(cc)}\in \R^{L\times L}
\end{split}
\end{equation}
and 
\begin{equation}
    \begin{split}
        [[\xi_2]_{cc'}]_{l l'} &=  \left( -\frac{1}{4}(v^h_{l} - v^h_{l'})^T\beta^{-1}(v^h_{l} - v^h_{l'})\right)\sum_{j = 1}^M  \sigma^4_{f_c} exp\left((\lambda_{j} - v^{\lambda}_{lc})^T \gamma^{-1} (\lambda_{j} - v^{\lambda}_{lc}) \right) exp\left((\lambda_{j} - v^{\lambda}_{l'c})^T \gamma^{-1} (\lambda_{j} - v^{\lambda}_{l'c}) \right)
        \\ & \Biggl(\sum_{i=1}^N y_{ij} r_{ic} r_{ic'} \sqrt{\frac{\det(\beta)}{\det\left(\beta + 2 \Sigma^h_{i}\right)}} exp\left(-\frac{1}{2} \left(\mu^h_{i} - \frac{v^h_{lc} - v^h_{l'c}}{2}\right)^T \left(\beta + 2\Sigma^h_{i}\right)^{-1} \left(\mu^h_{i} - \frac{v^h_{lc} - v^h_{l'c}}{2}\right) \right)
        \\ & + \sum _{i=1}^{N^*} y^*_{ij} \left(\mu^r_{ic} \mu^r_{ic'} + [\Sigma^r_{i\cdot}]_{cc'} \right) \sqrt{\frac{\det(\beta)}{\det\left(\beta + 2\Sigma^{x^*}_{i}\right)}} 
        \\ &  \quad exp\left(-\frac{1}{2} \left(\mu^{x^*}_{i} - \frac{v^h_{lc} - v^h_{l'c}}{2}\right)^T \left(\beta + 2\Sigma^{x^*}_{i}\right)^{-1} \left(\mu^{x^*}_{i} - \frac{v^h_{lc} - v^h_{l'c}}{2}\right) \right) \Biggr)
    \end{split}
\end{equation}

\section{Code, Data and Computation}
\label{appendix: reproducability}
\subsection{Code}

The code for this work provided is provided in the Supplementary Material. The code is provided with the seeds used to produce the figures and results here. 
The data for each run is also provided in a supplementary material. 

\subsection{Packages}

The code for WS-GPLVM was implemented using PyTorch version 2.1 \citep{paszke2019pytorch}. The comparison methods were implemented using GpyTorch version 1.11 \citep{gardner2018gpytorch} for the Gaussian Processes, and Scikit Learn version 1.2 \citep{pedregosa2011scikit} for PLS.

\subsection{Data Availability}

All the results for experiments in Section~\ref{sec:experiments} are from public datasets:
\begin{itemize}
    \item The data for the varying temperate spectroscopy was initially introduced by \cite{wulfert1998influence}. This dataset was  provided by the Biosystems Data Analysis Group at the Universiteit van Amsterdam. The dataset is governed by a specific license that restricts its use to research purposes only and prohibits any commercial use. All copyright notices and the license agreement have been adhered to in any redistribution of the dataset or its modifications. The original creators, the Biosystems Data Analysis Group of the Universiteit van Amsterdam, retain all ownership rights to the dataset. The data is available from \url{https://github.com/salvadorgarciamunoz/eiot/tree/master/pyEIOT} \cite{munoz2020supervised}, under the temperature effects example. 
    \item The oil flow data \citep{bishop1993analysis} is available from \url{http://staffwww.dcs.shef.ac.uk/people/N.Lawrence/resources/3PhData.tar.gz}.
    \item The remote sensing example comes from the UCR time series database \citep{ baldridge2009aster, dau2019ucr} and is available from \url{https://www.timeseriesclassification.com/aeon-toolkit/Rock.zip}.
\end{itemize}

\subsection{Computation}

All models can be run on an M1 Macbook Pro with 16GB of ram. To allow for parallelization of many experiments and restarts, some experiments were run on CPUs with 32GB RAM and some were run using cuda on Nvidia 4090 GPUs.


\section{Baseline Methods}
\label{appendix:baseline_details}

\subsection{Inverse Linear Model of Coregionalization (ILMC)}

While the true data generating process that we consider in spectroscopy is one in which the observations are a function of the weights, it is also possible to predict the weights using an inverse model, such as
\begin{equation}
    r_{i \cdot} = g(y_{i\cdot}) + \eta_{i},
\end{equation}
where $\eta_i \in \R^C$ is some noise. As this model reverses causality, in that the weights are a function of the data rather than the data being a function of the weights, we refer to it as an inverse model. 

As we are comparing to Bayesian methodologies we model $g(\cdot)$ with a Gaussian Process. As $r_{i\cdot}$ outputs covary with one another we use a linear model of coregionalization model. Additionally, we used a sparse GP \citep{hensman2013gaussian} for computational speed. 

For datasets with high input dimensions, \textit{i.e.} the spectroscopy regression and classification examples, principal component analysis (PCA) is applied to $Y$ and $Y^*$ before fitting the model, similar to \citep{chen2007gaussian}. The Gaussian process then takes the PCA reduced data, $y^{PCA}_{i\cdot} \in \mathbb{R}^{N_{\text{PCA}}}$ as input
\begin{equation}
    r_{i \cdot} = g(y^{\text{PCA}}_{i\cdot}) + \eta_{i},
\end{equation}

where $N_{\text{PCA}} < N$. We selected the number of principal components, $N_{\text{PCA}}$, by fitting multiple ILMC models to PCA-reduced data with different dimensions and selecting the one with the highest ELBO. 

For classification the prediction from the LMC was put through a Softmax Likelihood and a sparse GP variational inference was used to make predictions \citep{hensman2015scalable}.

\subsection{Classical Least Squares(CLS)/Classical Least Squares Gaussian Process (CLS-GP)}

The CLS/CLS-GP baseline is equivalent to the WS-GPLVM without the latent variables. CLS corresponds to the WS-GPLVM-ind. with no covariance between wavelengths, while CLS-GP includes a GP prior across the wavelengths. 

We treat this model in a Bayesian way, with the same priors as the WS-GPLVM/WS-GPLVM-ind., with the omission of the latent variables. 
This model does not have a conjugate posterior and we use a variational approximation, using a variational posterior that 
\begin{equation}
    q(F) = \prod_{j=1}^M q(f_i) \quad \text{where} \quad  q(f_{j\cdot}) = N(\mu^f_{j}, \Sigma^f_j). 
\end{equation}
Note that because the pure signals do not vary with a latent variable in this model, they are equal across all data points, so we do not need to index over $i$. With this distribution and the likelihood in Eq.~\eqref{eq: data generation} the elbo can be found using 
\begin{equation}
    \mathcal{L} = \langle d(Y| F,R^*,R)\rangle_{q(F,R^*)} - D_{KL}(q(F)|| p(F)) - D_{KL}(q(R^*)|| p(R^*)) + log(p(R)).
\end{equation}
The variational parameters (Dirichlet concentration parameter $\alpha_i$, $\mu^f_{j}$ and  $\Sigma^f_j$), along with the GP parameters for CLS-GP, are optimized numerically.

\subsection{Partial Least Squares (PLS)}

PLS is the standard tool for spectroscopy from chemometrics \citep{wold2001pls}. This method uses the heuristic of maximising covariance between the observed spectra and the chemical concentrations to select a linear latent space, based on which a linear predictive model is used. For classification PLS-Discriminant Analysis is used \citep{barker2003partial}.

\section{Implementation Details} \label{appendix: Implementation Details}

\subsection{Weighted Sum Gaussian Process Latent Variable Model}

\paragraph{Kernel} We use a squared exponential kernel:

\begin{equation}
\begin{split}
    k_c(&(h_i, \lambda_j),  (h_{i'}, \lambda_{j'})) = k((h_i, \lambda_j),  (h_{i'}, \lambda_{j'})) = 
    \\ & \sigma^2_{f} exp\left(-\frac{1}{2}\left(h_i - h_{i'}\right)^T \beta^{-1} \left(h_i - h_{i'}\right)\right) 
    \\ & exp\left(-\frac{1}{2}\left(\lambda_j - \lambda_{j'}\right)^T \gamma^{-1}\left(\lambda_j - \lambda_{j'}\right)\right) \quad \forall \; c,
\end{split}
\label{equation:ws-gplvm kernel}
\end{equation}

for the WS-GPLVM and WS-GPLVM-ind due ARD properties and analytical computation of the kernel expectations. 

\paragraph{Inducing Points}

We define our inducing points $v^h_{k} \in \R^A$ and $v^\lambda_{k'} \in \R$ for the latent and observed space respectively. Our inducing points for each component are then:
\begin{equation}
    V_c = V = \{v^h_{k}, v^\lambda_{k'}\}_{k=1\dots L_h,  k'=1\dots L_\lambda} \; \forall \; c.
\end{equation}

\paragraph{Hyperparameter and Inducing Point Initialization} Table~\ref{tab:hyperparameter_inits} gives full details of the hyperparameter initialisations for the WS-GPLVM and WS-GPLVM-ind for the different experiments. As we do not standardize the input (i.e. the wavelengths) the input lengthscale initialization is dependent on the range of the input. The latent variables are initialized to standard normal $h_i \sim \mathcal{N}(0, 1)$ prior, and the weights are initialized to the Dirichlet prior $r_{i\cdot} \sim Dir(1_c)$ for the regression settings. For the classification case, the weights are initialized to the prior, which is a uniform prior over the unit vectors. 

\begin{table}[h]
    \centering
    \begin{tabular}{|c|c|c|c|} \hline 
         Hyperparameter &  Spectroscopy  & Oil flow  & Spectroscopy \\
         & regression &  regression & classification \\
         \hline
         noise variance $\sigma^2$ & $1$ & $1$ & $1$\\ 
         kernel variance $\sigma_f^2$ & \text{Uniform}(0.5, 1) & \text{Uniform}(0.5, 1)& \text{Uniform}(0.5, 1) \\ 
         latent lengthscale $\beta$ &  \text{Uniform}(0.5, 5.5) & \text{Uniform}(0.1,1.1) & \text{Uniform}(0.1,1.1) \\ 
         latent inducing points $V^{h}$ & \text{Normal}(0, 1) & \text{Normal}(0, 1) & \text{Normal}(0, 1)\\ 
         number of latent inducing points $L_\lambda$ & 6 & 50 & 16\\ 
         input lengthscale $\gamma$ & 1000 & n/a & 40000\\ 
         input inducing points $V^{\lambda}$  & grid  & n/a & grid \\ 
         number of input inducing points $L_h$ & 28 & n/a & 20 \\ 
         latent dimensions $A$ & 2 & 5 & 5 \\ \hline
         \end{tabular}
    \caption{Hyperparameter initializations for WS-GPLVM and WS-GPLVM-ind. When the initialization is a distribution, the initial value is drawn from that distribution. Initalization is 'grid' means a grid of points across the whole input domain. }
    \label{tab:hyperparameter_inits}
\end{table}

\paragraph{Optimization} As described in Section~\ref{section: implimentation}, the optimization of the WS-GPLVM is done in three steps. (1) The hyperparameters are trained on the training data without allowing $h$ to train using Adam \citep{kingma2014adam} with a learning rate of $0.05$. (2) The noise variance is attenuated over $20$ steps from $1$ to the value it had trained to in the previous stage. At each stage, two steps of the Limited-memory Broyden-Fletcher-Goldfarb-Shanno (L-BFGS) optimizer are taken. (3) All parameters are trained together using L-BFGS. 

\paragraph{Random Restarts} Optimizing the ELBO with respect to GP hyperparameters often encounters local optima \cite[Ch.~5]{rasmussen2003gaussian}, making the results sensitive to initialization. To address this, we use random restarts with varied initializations, selecting the run that yields the highest ELBO.

Increasing the number of restarts can either improve the final ELBO or maintain the current best if the optimal solution has already been found. To determine an appropriate number of restarts for the WS-GPLVM, we conducted additional trials: 20 restarts for the oil flow experiment and 32 for the remote sensing rock classification. While it can't be proven if the global optimum has been reached, the ELBO plateaued after 15 restarts for the oil flow data and after just 5 for the rock classification, indicating that a few restarts are typically sufficient to reach a good local optimum.

\subsection{Inverse Linear model of Coregionalization
}

\paragraph{Kernel} The ILMC uses the squared exponential kernel: 
\begin{equation}
k(\lambda_j, \lambda_{j'}) = \sigma^2_f \exp\left({-\frac{1}{2}(\lambda_j - \lambda{j'})^T \gamma^{-1}(\lambda_j - \lambda{j'})}\right).
\label{eq:squarey_exp}
\end{equation}

\paragraph{Hyperparameter and Inducing Point Initialization} For the ILMC baseline, the lengthscales $\gamma$ and kernel variances $\sigma^2_f$ are initialized to one. $50$ inducing points are initialized as $v_\lambda \sim \text{Normal}(0,1)$. 

\paragraph{Optimization} We use Adam \citep{kingma2014adam} to optimize the ILMC hyperparameters. We use a learning rate of $0.01$ for the spectroscopy regression and classification experiments, and $0.1$ for the oil flow experiment.

\subsection{Classical Least Squares Gaussian Process}

The Classical Least Squares Gaussian process corresponds to the WS-GPLVM model without the latent space to allow the pure component signals to vary. 

\paragraph{Kernel} The CLS-GP uses the squared exponential kernel, shown in Equation~\ref{eq:squarey_exp} for the pure component signals.

\paragraph{Hyperparameter and Inducing Point Initialization} The CLS-GP hyperparameters are initialized in the same way as the WS-GPLVM, shown in Table~\ref{tab:hyperparameter_inits}, except the CLS-GP has no latent lengthscale, dimensions or inducing points. The weights are initialized in the same way. 

\paragraph{Optimization} All the CLS parameters are optimized together using L-BFGS. 

\subsection{Classical Least Squares}

Classical Least Squares corresponds to the WS-GPLVM-ind. model without the latent space to allow the pure component signals to vary. 

\paragraph{Hyperparameter Initialization} For CLS, the weights were initialized to the prior as they are for the WS-GPLVM. The noise variance was initialized to $0.01$. 

\paragraph{Optimization} The weights and CLS parameters are optimized together using L-BFGS. 

\subsection{Partial Least Squares} 

We use the Scikit Learn version 1.2 \citep{pedregosa2011scikit} implementation for PLS. The only hyperparameter in PLS is the number of components. We select the number of components by running a 10-fold cross validation, or two-fold for the oil flow dataset, on the training set and selecting the one with the lowest mean squared error.

\section{Additional Results} \label{appendix:additional_results}

Table~\ref{tab:regression results} and\ref{tab:classification results} show the mean and standard deviation of the performance metrics for the two regression and one classification experiments. We include the WS-GPLVM-ind for the near infra-red spectroscopy regression and remote sensing rock classification examples for comparison. For both these examples, the WS-GPLVM-ind performs worse than the WS-GPLVM justifying the inclusion of the dependence on wavelengths for these examples. However, it still outperforms the ILMC and CLS-GP, indicating it can still reasonably predict the weights using only the latent variables.

\begin{table}[ht!]
\centering
\begin{tabular}{|l||c|c||c|c||}
\hline
 & \multicolumn{2 }{c||} {\textit{Spectroscopy}} & \multicolumn{2 }{c||} {\textit{Oil Flow}}\\\hline
\textbf{Method} & \textbf{NLPD)} & \textbf{MSE} {\tiny$[10^{-4}]$} & \textbf{NLPD (Oil Flow)} & \textbf{MSE } {\tiny$[10^{-4}]$} \\
\hline
WS-GPLVM & -561(9.22) & \textbf{2.67(0.78)} & - & - \\
\hline
WS-GPLVM-ind. & -516(57.5) & 58.4(45.8) & -7430(81.12) & 5.97(4.58) \\
\hline
CLS / CLS-GP & \textbf{-693(7.91)} & 124(16.2) & -3508(24.6) & 81.5(8.2) \\
\hline
ILMC & -309(3.21) & 505(25.5) & \textbf{-8228(112)} & \textbf{2.82(0.42)} \\
\hline
PLS & n/a & 3.04(0.73) & n/a & 3.81(0.13) \\
\hline
\end{tabular}
\caption{Results of the Weighted-Sum GPLVM (WS-GPLVM), Multi-output Gaussian Process (ILMC), and Partial Least Squares (PLS). Regression results show Negative Log Predictive Density (NLPD) and  Mean Squared Error (MSE) of the prediction. Each example was run 10 times and the mean is quoted here, along with two times the standard error of the mean in brackets. For results that deviate significantly from one, powers of ten were removed from all of the results and where applicable these are shown in square brackets at the top of the column. The best mean for each column is in bold.}
\label{tab:regression results}
\end{table}

\begin{table}[ht!]
\centering
\begin{tabular}{|l||c|c|c||}
\hline
& \multicolumn{3 }{c||} {\textit{Remote Sensing}} \\
\hline
\textbf{Method} & \textbf{LPP} & \textbf{Acc.} {\tiny$[10^{-2}]$} & \textbf{AUC} \\
\hline
WS-GPLVM & -14.2(17.1) & 81.4(21.3) & \textbf{0.91(0.13)} \\
\hline
WS-GPLVM-ind. & -21.0(17.5) & 75.7(17.9) & 0.89(0.12) \\
\hline
CLS / CLS-GP & -53.7(27.8) & 60.0(14.8) & 0.81(0.11) \\
\hline
ILMC & \textbf{-9.27(2.35)} & 30.0(12.5) & 0.68(0.18) \\
\hline
PLS & n/a & \textbf{82.9(16.2)} & n/a \\
\hline
\end{tabular}
\caption{Results of the Weighted-Sum GPLVM (WS-GPLVM), Multi-output Gaussian Process (ILMC), and Partial Least Squares (PLS). Classification results report the Log Predicted Probability (LPP), the Accuracy (Acc.) and the Area Under the Receiver Operator Curve (AUC). Each example was run 10 times and the mean is quoted here, along with two times the standard error of the mean in brackets. For results that deviate significantly from one, powers of ten were removed from all of the results and where applicable these are shown in square brackets at the top of the column. The best mean for each column is in bold.}
\label{tab:classification results}
\end{table}

\section{Additional Details for Experiment Setup}
\label{appendix: additional details}

\subsection{Toy Example}
For this data set, we used a 5 dimensional latent for the WS-GPLVM with a grid of inducing points with 8 inducing point locations in the variable space and 16 inducing points in the observed space, for a total of 128 inducing points. 

\subsection{Near Infra Red Spectroscopy with varying temperatures}

The original near infra red spectroscopy dataset contained a region of constant signal at low wavelengths was wel as very noisy signals at the highest wavelengths. We removed both these regions before the pre-processing and analysis steps, truncating the wavelengths to be between 800-1000$nm$. For this experiments we used 2 latent dimensions, 6 latent inducing points and 28 input inducing points.  

\subsection{Oil Flow Regression}
This example was run with independent measurement locations. For the WS-GPLVM, we used 50 inducing points and 5 latent dimensions. 


\subsection{Remote Sensing Rock Classification}

\begin{figure}
    \centering
    \includegraphics[width=0.6\textwidth]{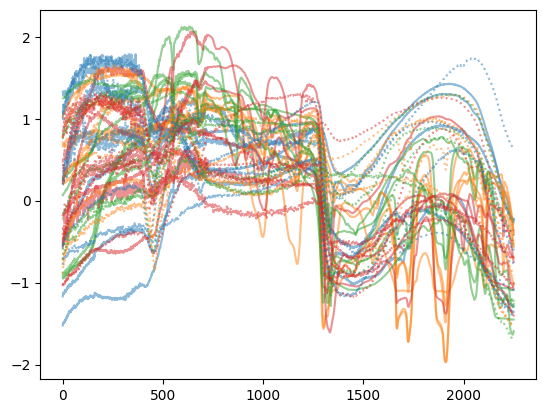}
    \caption{Training and test data for the remote sensing rock classification example.}
    \label{fig:spectra rocks}
\end{figure}

Similar to the near infra-red spectroscopy regression example, the largest wavelengths ($>2250nm$) were noisy so we removed them from the data. For the WS-GPLVM we used 5 latent dimensions, 16 latent inducing points and 20 input inducing points. We found this number of inducing points to give a reasonable trade-off between computational time and quality of the ELBO optimization. 

\end{document}